\documentclass[runningheads]{llncs}

\usepackage[utf8]{inputenc}
\usepackage[OT1]{fontenc}

\usepackage{graphicx}
\usepackage[font=small,labelsep=period,labelfont=bf,belowskip=0pt,aboveskip=4pt,compatibility=false]{caption}
\usepackage{subcaption}
\usepackage{amsmath,amssymb} %
\usepackage[table,usenames,dvipsnames]{xcolor}
\usepackage[width=122mm,left=12mm,paperwidth=146mm,height=193mm,top=12mm,paperheight=217mm]{geometry}

\usepackage{tikz}
\usetikzlibrary{backgrounds,calc,external,spy}

\tikzset{annotate/.style={ %
execute at begin scope={
	\pgfkeys{src/.store in=\src,src/.value required}
	\pgfkeys{pxwidth/.store in=\pxwidth,pxwidth/.value required}
	\pgfkeys{#1}
	\pgftransformshift{\pgfpointanchor{\src}{north west}}
	\pgfsetyvec{\pgfpointscale{1/\pxwidth}{\pgfpointdiff{\pgfpointanchor{\src}{north west}}{\pgfpointanchor{\src}{south west}}}}
	\pgfsetxvec{\pgfpointscale{1/\pxwidth}{\pgfpointdiff{\pgfpointanchor{\src}{north west}}{\pgfpointanchor{\src}{north east}}}}
}}}

\usepackage[capitalise]{cleveref}
\Crefname{equation}{Equation}{Equations}
\Crefname{figure}{Figure}{Figures}
\crefformat{equation}{Equation~#2#1#3}
\crefrangeformat{equation}{Equations~#3#1#4 to~#5#2#6}
\crefmultiformat{equation}{Equations~#2#1#3}{ and~#2#1#3}{, #2#1#3}{ and~#2#1#3}
\usepackage{multirow}
\usepackage{colortbl}
\usepackage{enumitem}

\usepackage{array}
\newcolumntype{?}{!{\vrule width 1pt}}
\newcolumntype{C}[1]{>{\centering\arraybackslash}p{#1}}

\newlength{\Oldarrayrulewidth}
\newcommand{\Cline}[2]{%
	\noalign{\global\setlength{\Oldarrayrulewidth}{\arrayrulewidth}}%
	\noalign{\global\setlength{\arrayrulewidth}{#1}}\cline{#2}%
	\noalign{\global\setlength{\arrayrulewidth}{\Oldarrayrulewidth}}}

\usepackage{floatrow}
\newfloatcommand{capbtabbox}{table}[][\FBwidth]
\newcommand{\HR}[1]{\textcolor{magenta}{[\textbf{Helge}: #1]}}

\newcommand{\CR}[1]{\textcolor{Cerulean}{[\textbf{CR}: #1]}}

\newcommand{\NEW}[1]{#1}
\newcommand{\inlineheading}[1]{\vspace{0.5em}\noindent\textbf{#1.\hspace{0.5em}}}
\newcommand{\IGNORE}[1]{{}}

\newcommand{\detection}{\mathcal{D}}
\newcommand{\skel}{\mathcal{K}}
\newcommand{\Poseset}{\mathbf{P}}
\newcommand{\pose}{\mathbf{p}}
\newcommand{\bones}{\mathbf{b}}

\newcommand{\volume}{\boldsymbol{\gamma}}
\newcommand{\shape}{\mathbf{s}}
\newcommand{\I}{\mathcal{I}}
\newcommand{\etal}{et al.}
\newcommand{\ie}{i.e.~}

\newcommand{\B}{\mathcal{B}}

\newcommand{\norm}[1]{\left\lVert#1\right\rVert}

\newcommand{\s}{{\sigma}}
\newcommand{\cam}{\mathbf{o}}

\newcommand{\n}{\mathbf{n}}

\newcommand{\V}{\mathcal{V}}
\newcommand{\sR}{{\bar{\sigma}}}
\newcommand{\cR}{{\bar{c}}}

\newcommand{\m}{\boldsymbol{\mu}}

\newcommand*\dif{\mathop{}\!\mathrm{d}} %

\newcommand{\dt}{\dif t}
\newcommand{\ds}{\dif s}

\newcommand{\IG}{\nabla \I}
\newcommand{\BG}{\nabla \B}
\newcommand{\nG}{\nabla \n}

\begin{document}

\pagestyle{headings}
\mainmatter

\title{General Automatic Human Shape and Motion Capture Using Volumetric Contour Cues}

\titlerunning{General Automatic Human Shape and Motion Capture}

\authorrunning{Rhodin \etal}

\author{%
Helge Rhodin$^\text{1}$\quad%
Nadia Robertini$^\text{1, 2}$\quad%
Dan Casas$^\text{1}$\\
Christian Richardt$^\text{1, 2}$\quad%
Hans-Peter Seidel$^\text{1}$\quad%
Christian Theobalt$^\text{1}$%
}
\institute{
$^\text{1}$MPI Informatik,
$^\text{2}$Intel Visual Computing Institute
}

\maketitle

\begin{abstract}
Markerless motion capture algorithms require a 3D body with properly personalized skeleton dimension and/or body shape and appearance to successfully track a person.
Unfortunately, many tracking methods consider model personalization a different problem and use manual or semi-automatic model initialization, which greatly reduces applicability.
In this paper, we propose a fully automatic algorithm that jointly creates a rigged actor model commonly used for animation -- skeleton, volumetric shape, appearance, and optionally a body surface -- and estimates the actor's motion from multi-view video input only.
The approach is rigorously designed to work on footage of general outdoor scenes recorded with very few cameras and without background subtraction.
Our method uses a new image formation model with analytic visibility and analytically differentiable alignment energy.
For reconstruction, 3D body shape is approximated as a Gaussian density field.
For pose and shape estimation,
we minimize a new edge-based alignment energy inspired by volume ray casting in an absorbing medium.
We further propose a new statistical human body model that represents the body surface, volumetric Gaussian density, and variability in skeleton shape.
Given any multi-view sequence, our method jointly optimizes the pose and shape parameters of this model fully automatically in a spatiotemporal way.

\end{abstract}
\section{Introduction}

Markerless full-body motion capture techniques refrain from markers used in most commercial solutions, and promise to be an important enabling technique in 
computer animation and visual effects production, in sports and biomechanics research, and the growing fields of virtual and augmented reality.
While early markerless methods were confined to indoor use in more controlled scenes and backgrounds recorded with eight or more cameras~\cite{moeslund2006survey}, recent methods succeed in general outdoor scenes with much fewer cameras~\cite{moeslund2012,elhayek2015efficient}.
Before motion capture commences, the 3D body model for tracking needs to be personalized to the captured human. %
This includes personalization of the bone lengths, but often also of biomechanical shape and surface, including appearance.
This essential initialization is, unfortunately, neglected by many methods and solved with an entirely different approach, or with specific and complex manual or semi-automatic initialization steps.  
For instance, some methods for motion capture in studios with controlled backgrounds rely on static full-body scans~\cite{de2008performance,gall2009motion,zollhofer2014real}, or personalization on manually segmented initialization poses~\cite{stoll2011iccv}.
Recent outdoor motion capture methods use entirely manual model initialization~\cite{elhayek2015efficient}.
When using depth cameras, automatic model initialization was shown \cite{shotton2013acm,bogo2015detailed,tong2012tvcg,Helten:2013-3DV,newcombe2015dynamicfusion}, but RGB-D cameras are less accessible and not usable outdoors.
Simultaneous pose and shape estimation from in-studio multi-view footage with background subtraction was also shown~\cite{kakadiaris98,ahmed2005automatic,Balan:2007}, but not on footage of less constrained setups such as outdoor scenes filmed with very few cameras. 
We therefore propose a fully-automatic space-time approach for simultaneous model initialization and motion capture.
Our approach is specifically designed to solve this problem automatically for multi-view video footage recorded in general environments (moving background, no background subtraction) and filmed with as few as two cameras.
Motions can be arbitrary and unchoreographed.
It takes a further step towards making markerless motion capture practical in the aforementioned application areas, and enables motion capture from third-party video footage, where dedicated initialization pose images or the shape model altogether are unavailable.
Our approach builds on the following contributions.

\begin{figure}[t]
	\centering%
	\resizebox{\columnwidth}{!}{%
		\input{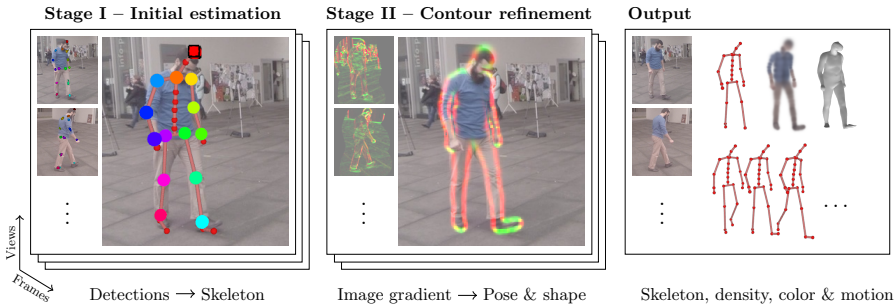}%
	}%
	\caption{\label{fig:overview}%
		Method overview. Pose is estimated from detections in Stage~I, actor shape and pose is refined through contour alignment in Stage~II by space-time optimization. Outputs are the actor skeleton, attached density, mesh and motion.}
\end{figure}

First, we introduce a body representation that extends a scene model 
inspired by light transport in absorbing transparent media~\cite{rhodin2015iccv}.
We represent the volumetric body shape by Gaussian density functions attached to a kinematic skeleton.
We further define a novel 2D contour-based energy that measures contour alignment with image gradients on the raw RGB images using a new volume raycasting image formation model.
We define contour direction and magnitude for each image position, which form a ridge at the model outline, see \cref{fig:overview}.
No explicit background segmentation is needed. 
Importantly, our energy features analytic derivatives, including fully-differentiable visibility everywhere. %

The second contribution is a new data-driven body model that represents human surface variation, the space of skeleton dimensions, and the space of 
volumetric density distributions optimally for reconstruction using a low-dimensional parameter space. 

\noindent
Finally, we propose a space-time optimization approach that fully automatically computes both the shape and the 3D skeletal pose of the actor using both contour and ConvNet-based joint detection cues.
The final outputs are (1) a rigged character, as commonly used in animation, comprising a personalized skeleton and attached surface, along with the (optionally colored) volumetric human shape representation, and (2) the joint angles for each video frame.
We tested our method on eleven sequences, indoor and outdoor, showing reconstructions with fewer cameras and less manual effort compared to the state of the art.

\section{Related work}
\label{sec:relatedWork}

Our goal is to fully automatically capture a personalized, rigged surface model, as used in animation, \emph{together with} its sequence of skeletal poses from sparse multi-view video of general scenes where background segmentation is hard.
Many multi-view markerless motion capture approaches consider model initialization and tracking separate problems~\cite{moeslund2012}.
Even in recent methods working outdoors, shape and skeleton dimensions of the tracked model are either initialized manually prior to tracking~\cite{elhayek2015efficient}, or estimated from manually segmented initialization poses~\cite{stoll2011iccv}.
In controlled studios, static shape \cite{hilton1999virtual,bualan2008naked} or dimensions and pose of simple parametric human models \cite{wren1997pfinder,kakadiaris98,hilton2000whole,mikic2003human,ahmed2005automatic} can be optimized by matching against chroma-keyed multi-view image silhouettes.
Many multi-view performance capture methods~\cite{Theobalt2010} deform a static full-body shape template obtained with a full-body scanner~\cite{de2008performance,gall2009motion,wu2013onset,wu2012full}, or through fitting against the visual hull~\cite{Vlasic2008,starck2003model,ballan2008marker,allain2015efficient} to match scene motion.
Again, all these require controlled in-studio footage, an off-line scan, or both.
Shape estimation of a parametric model in single images using shading and edge cues~\cite{guan2009estimating}, 
actor silhouettes \cite{sminchisescu2002human,grest2005human,sigal2007combined,chen2010inferring,zhou2010parametric},
or monocular pose and shape estimation from video is also feasible \cite{Guo2012,hasler2010multilinear,Jain:MovieReshape}, but require substantial manual intervention (joint labeling, feature/pose correction, background subtraction etc.).
For multi-view in-studio setups (3–4 views), where background subtraction works, B{\u{a}}lan \etal~\cite{Balan:2007} estimate shape and pose of the SCAPE parametric body model and Cheung et al.~propose a model-free approach \cite{cheung2003shape}.
Optimization is independent for each frame and requires initialization by a coarse cylindrical shape model.
\NEW{
Implicit surface representations yield beneficial properties for pose \cite{plankers2003articulated} and surface \cite{ilic2006implicit} reconstruction, but do not avoid the dependency on explicit silhouette input.}
In contrast to all previously mentioned methods, our approach requires no manual interaction, succeeds even with only two camera views, and on scenes recorded outdoors without any background segmentation.

Recently, several methods to capture both shape and pose of a parametric human body model with depth cameras were proposed~\cite{bogo2015detailed,Helten:2013-3DV,Cui2013}; these special cameras are not 
as easily available and often do not work outdoors. We also build up on the success of parametric body models for surface representation, e.g.~\cite{Jain:MovieReshape,Anguelov2005,pishchulin2015building,loper2014mosh}, but extend these models to represent the space of volumetric shape models needed for tracking, along with a rigged surface and skeleton. 

Our approach is designed to work without explicit background subtraction.
In outdoor settings with moving backgrounds and uncontrolled illumination, such segmentation is hard, but progress has been made by multi-view segmentation \cite{CampbVHC2007,wang2014wide,djelouah2015sparse}, joint segmentation and reconstruction \cite{Szeliski1998,guillemaut2011joint,Bray:2006,Mustafa2015}, and also aided by propagation of a manual initialization \cite{hasler2009cvpr,wu2013onset}.
However, the obtained segmentations are still noisy, enabling only rather coarse 3D reconstructions~\cite{Mustafa2015}, and many methods would not work with only two cameras.

Edge cues have been widely used in human shape and motion estimation \cite{moeslund2006survey,moeslund2012,deutscher2000articulated,sidenbladh2003learning,sigal2012loose,kehl2005markerless}, but we provide a new formulation for their use and make edges in general scenes the primary cue.
In contrast, existing shape estimation works use edges are supplemental information, for example to find self-occluding edges in silhouette-based methods and to correct rough silhouette borders \cite{guan2009estimating}.
Our new formulation is inspired by the work of Nagel \etal, where model contours are directly matched to image edges for rigid object \cite{kollnig19953d} and human pose tracking \cite{wachter1997tracking}.
Contour edges on tracked meshes are found 
by a visibility test, and are convolved with a Gaussian kernel.
This approach forms piecewise-smooth and differentiable model contours which are optimized to maximize overlap with image gradients.
\NEW{
The work of Sminchisescu and Triggs \cite{sminchisescu2003estimating} is closely related, an actor model consisting of superquadric ellipsoids is used for monocular 3D pose tracking and is initialized by optimizing bone-length and ellipsoid deformation parameters with respect to image edge cues. However, manual annotation of skeleton joint locations and coarse pose initialization is required.
}
We advance existing approaches in several ways: our model is volumetric, analytic visibility is incorporated in the model and optimization, occlusion changes are differentiable, the human is represented as a deformable object, allowing for shape estimation, and contour direction is handled separately from contour magnitude.

Our approach follows the generative analysis-by-synthesis approach: contours are formed by a 3D volumetric model and image formation is an extension of the volumetric ray-tracing model proposed by Rhodin et al. \cite{rhodin2015iccv}. Many discriminative methods for 2D pose estimation were proposed~\cite{tompson2014joint,felzenszwalb2005pictorial,andriluka2009pictorial}; multi-view extensions were also investigated \cite{sigal2012loose,amin2013multi,belagiannis20143d}.
Their goal is different to ours, as they find single-shot 2D/3D joint locations, but no 3D rigged body shape and no temporally stable joint angles needed for animation. 
We thus use a discriminative detector only for initialization.  
Our work has links to non-rigid structure-from-motion that finds sparse 3D point trajectories (e.g. on the body) from single-view images of a non-rigidly moving scene~\cite{Park2015}.
Articulation constraints~\cite{fayad2011automated} can help to find the sparse scene structure, but the goal is different from our estimation of a fully dense, rigged 3D character and stable skeleton motion.  
\section{Notation and overview}
\label{sec:overview}

Input to our algorithm are RGB image sequences $\I_{c,t}$, recorded with calibrated cameras $c \!=\! 1,\ldots,C$ and synchronized to within a frame (see list of datasets in supplemental document).
The output of our approach is the configuration of a virtual actor model $\skel(\pose_t,\bones,\volume)$ for each frame $t \!=\! 1,\ldots,T$, comprising the per-frame joint angles $\pose_t$, the personalized bone lengths $\bones$, as well as the personalized volumetric Gaussian representation $\volume$, including color, of the actor.

In the following, we first explain the basis of our new image formation model, the Gaussian density scene representation, and our new parametric human shape model building on it (\cref{sec:bodyModel}).
Subsequently, we detail our space-time optimization approach (\cref{sec:objectiveEnergy}) in two stages:
(I) using ConvNet-based joint detection constraints (\cref{sec:detectionFit}); and
(II) using a new ray-casting-based volumetric image formation model and a new contour-based alignment energy (\cref{sec:contours}).

\IGNORE{
\begin{enumerate}[topsep=0.25em,labelindent=1em,leftmargin=*,label=\Roman*.]
\item using ConvNet-based joint detection constraints (\cref{sec:detectionFit}),
\item using a new ray-casting-based volumetric image formation model and
a new contour-based alignment energy (\cref{sec:contours}), and
\item appearance estimation and outlier rejection (\cref{sec:appearance}).
\end{enumerate}
}

\IGNORE{
The paper is structured as follows:
In Sec. \ref{sec:bodyModel} we explain our volumetric shape model and introduce a statistical body model that predicts and correlates $\bones,\volume$ through low-dimensional shape parameters, $\shape$. 
We optimize a spatio-temporal energy function by gradient descent to infer $\pose_t$ and $\shape$ (Sec. \ref{sec:objectiveEnergy}).
Optimization is started from 2D joint detections, which gives rough estimates of the subject's body parts in each video (Sec. \ref{sec:detections}).
The focus of this work is on estimating body shape, $\shape$, and to refine $\pose$ by fitting the contour of $\skel$ to the spatial gradients of the video input.
To this end we introduce an analytic contour model (Sec. \ref{sec:contours}) and corresponding contour similarity measure that operates on RGB input directly (Sec. \ref{sec:contourSimilarity}).
Further, we reject outliers to compensate for drastic failures of the initialization (Sec. \ref{sec:outlierRejection}).
Section \ref{sec:experiments} evaluates XXXX.
}
\section{Volumetric statistical body shape model}
\label{sec:bodyModel}

\NEW{
To model the human in 3D for reconstruction, we build on sum-of-Gaussians representations~\cite{stoll2011iccv,rhodin2015iccv} and model the volumetric extent of the actor using a set of 91 isotropic Gaussian density functions distributed in 3D space.}
Each Gaussian $G_q$ is parametrized by its standard deviation $\s_q$, mean location $\m_q$ in 3D, and density $c_q$, which define the Gaussian shape parameters $\volume \!=\! \{\m_q, \s_q, c_q\}_q$.
The combined density field of the Gaussians, $\sum_q c_q G_q$, smoothly describes the volumetric occupancy of the human in 3D space, see \cref{fig:overview}.
Each Gaussian is rigidly attached to one of the bones of an articulated skeleton with bone lengths $\bones$ and 16 joints, whose pose is parameterized with 43 twist pose parameters, \ie the Gaussian position $\m_q$ is relative to the attached bone.
This representation allows us to formulate a new alignment energy tailored to pose fitting in general scenes, featuring analytic derivatives 
and fully-differentiable visibility (\cref{sec:objectiveEnergy}).

In their original work, Rhodin et al. \cite{rhodin2015iccv} create 3D density models for tracked shapes by a semi-automatic placement of Gaussians in 3D.
Since the shape of humans varies drastically, a different distribution of Gaussians and skeleton dimensions is needed for each individual to ensure optimal tracking.
In this paper, we propose a method to automatically find such a skeleton and optimal attached Gaussian distribution, along with a good body surface.
Rather than optimizing in the combined high-dimensional space of skeleton dimensions, the number of Gaussians and all their parameters, we build a new specialized, low-dimensional parametric body model.

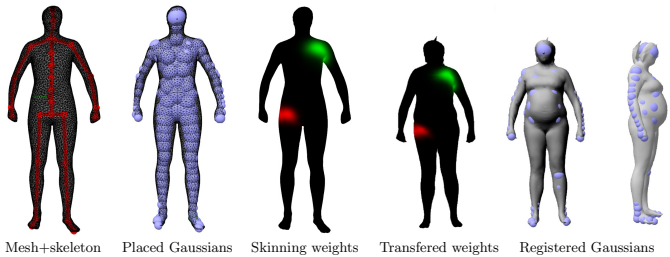
\begin{figure}[t!]
\centering%
\begin{tikzpicture}[tight background,label/.style={below=0mm, scale=0.6, align=center}]
\node[inner sep=0pt] (image) at (0,0) {\includegraphics[height=0.25\linewidth,trim=40px 10px 150px 0px,clip]{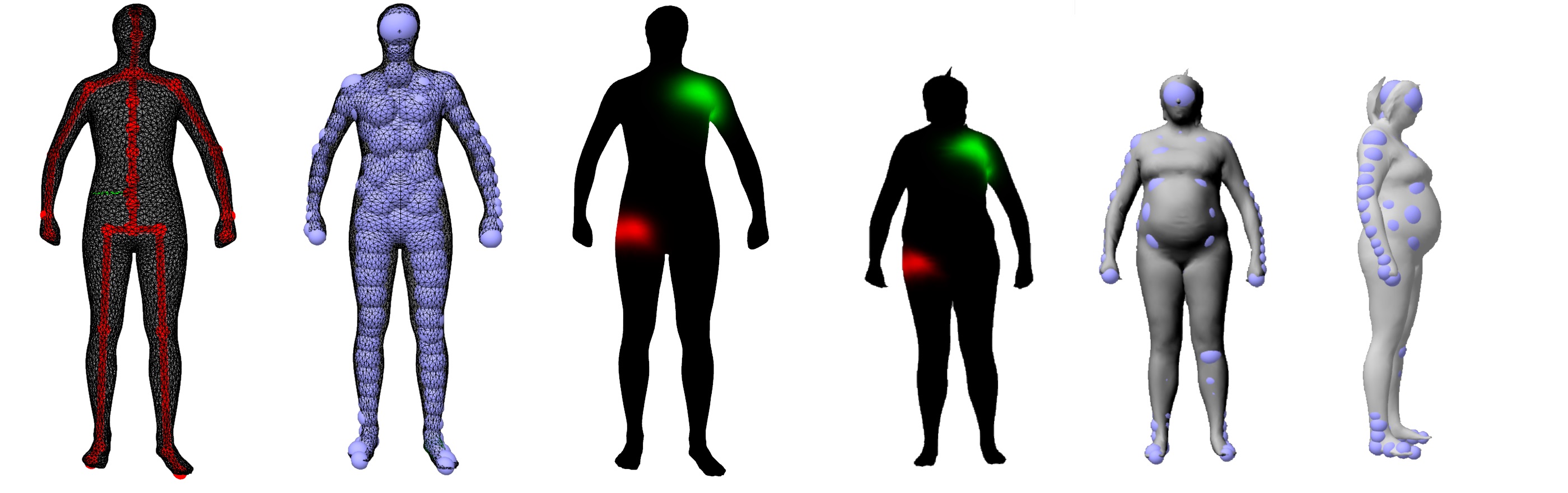}};

\begin{scope}[annotate={src=image,pxwidth=1}]
\node[label] (label1) at (0.075,1.0) {Mesh+skeleton};
\node[label] (label2) at (0.26, 1.0) {Placed Gaussians};
\node[label] (label3) at (0.45, 1.0) {Skinning weights};
\node[label] (label4) at (0.65, 1.0) {Transfered weights};
\node[label] (label5) at (0.87, 1.0) {Registered Gaussians};
\end{scope}

\end{tikzpicture}%
	\caption{
		Registration process of the body shape model. Skeleton and Gaussians are once manually placed into the \emph{reference} mesh, vertex correspondence transfers Gaussian- and joint-neighborhood weights (green and red respectively), to register reference bones and Gaussians to all instance meshes.}
	\label{fig:databaseCreation}
\end{figure}

Traditional statistical human body models represent variations in body surface only across individuals, as well as pose-dependent surface deformations using linear \cite{Allen2003,loper2015smpl} or non-linear \cite{Anguelov2005,hasler2010multilinear} subspaces of the mesh vertex positions.
For our task, we build an enriched statistical body model that parameterizes, in addition to the body surface, the optimal volumetric Gaussian density distribution $\volume$ for tracking, and the space of skeleton dimensions $\bones$, through linear functions $\volume(\shape)$, $\bones(\shape)$ of a low-dimensional shape vector $\shape$.
To build our model, we use an existing database of 228 registered scanned meshes of human bodies in neutral pose~\cite{hasler2009statistical}.
We take one of the scans as \emph{reference} mesh, and place the articulated skeleton inside.
The 91 Gaussians are attached to bones, their position is set to uniformly fill the mesh volume, and their standard deviation and density is set such that a density gradient forms at the mesh surface, see \cref{fig:databaseCreation} (left).
This manual step has to be done only once to obtain Gaussian parameters $\volume_\text{ref}$ for the database reference, and can also be automated by silhouette alignment \cite{rhodin2015iccv}.
The best positions $\{\m_q\}_q$ and scales $\{\s_q\}_q$ of Gaussians $\volume_i$ for each remaining database \emph{instance} mesh $i$ are automatically derived by weighted Procrustes alignment.
Each Gaussian $G_q$ in the reference has a set of neighboring surface mesh vertices.
The set is inferred by weighting vertices proportional to the density of $G_q$ at their position in the reference mesh, see \cref{fig:databaseCreation} (right). 
For each Gaussian $G_q$, vertices are optimally translated, rotated and  scaled to align to the corresponding instance mesh vertices.
These \IGNORE{affine}\NEW{similarity} transform\IGNORE{ation}s are applied on $\volume_\text{ref}$ to obtain $\volume_i$, where scaling multiplies $\s_q$ and translation shifts $\m_q$.

To infer the adapted skeleton dimensions $\bones_i$ for each instance mesh, we follow a similar strategy:
we place Gaussians of standard deviation 10\,cm at each joint in the reference mesh, 
which are then scaled and repositioned to fit the target mesh using the same Procrustes strategy as before.
This yields properly scaled bone lengths for each target mesh.

\NEW{
Having estimates of volume $\volume_i$ and bone lengths $\bones_i$ for each database entry~$i$, we now learn a joint body model.
We build a PCA model on the data matrix $\left[(\volume_1;\bones_1), (\volume_2;\bones_2), \ldots\right]$,
where each column vector $(\volume_i;\bones_i)$ is the stack of estimates for entry $i$.
The mean is the average human shape $(\bar{\volume};\bar{\bones})$, and the PCA basis vectors span the principal shape variations of the database.
The PCA coefficients are the elements of our shape model $\shape$, and hence define the volume $\volume(\shape)$ and bone lengths $\bones(\shape)$.
Due to the joined model, bone length and Gaussian parameters are correlated, and optimizing $\shape$ for bone length during pose estimation (stage~I) thus moves and scales the attached Gaussians accordingly.
To reduce dimensionality, we use only the first 50 coefficients in our experiments.
}

\NEW{
To infer the actor body surface, we introduce a volumetric skinning approach. 
The reference surface mesh is deformed in a free-form manner along with the Gaussian set under new pose and shape parameters.
Similar to linear blend skinning \cite{Lewis:2000}, each surface vertex is deformed with the set of 3D transforms of nearby Gaussians, weighted by the density weights used earlier for Procrustes alignment.
This coupling of body surfaces to volumetric model 
is as computationally efficient as using a linear PCA space on mesh vertices~\cite{pishchulin2015building}, while yielding comparable shape generalization and extrapolation qualities to methods using more expensive non-linear reconstruction~\cite{Anguelov2005}, see supplemental document.
Isosurface reconstruction using Marching Cubes would also be more costly~\cite{plankers2003articulated}.
}
\section{Pose and shape estimation}
\label{sec:objectiveEnergy}

We formulate the estimation of the time-independent $50$ shape parameters $\shape$ and the time-dependent $43T$ pose parameters $\Poseset \!=\! \{\pose_1,\ldots,\pose_T\}$ as a combined space-time optimization problem over all frames $\I_{c,t}$ and camera viewpoints $c$ of the input sequence of length $T$:
\begin{equation}
E (\Poseset\!,\shape) = E_\text{shape}(\shape) + \sum_{t} \!\Big( E_\text{smooth}(\Poseset\!,\shape,t) + E_\text{pose}(\pose_t) + \sum_{c} E_\text{data}(c,\pose_t,\shape) \Big)
\!\text{.}
\label{egn:objective}
\end{equation}
Our energy uses quadratic prior terms to regularize the solution:
$E_\text{shape}$ penalizes shape parameters that have larger absolute value than any of the database instances,
$E_\text{smooth}$ penalizes joint-angle accelerations to favor smooth motions, and
$E_\text{pose}$ penalizes violation of manually specified anatomical joint-angle limits.
The data term $E_\text{data}$ measures the alignment of the projected model to all video frames.
To make the optimization of \cref{egn:objective} succeed in unconstrained scenes with few cameras, we solve in two subsequent stages.
In Stage~I (\cref{sec:detectionFit}), we optimize for a coarse skeleton estimate and pose set without the volumetric distribution, but using 2D joint detections as primary constraints.
In Stage~II (\cref{sec:contours}), we refine this initial estimate and optimize for all shape and pose parameters using our new contour-based alignment energy.
Consequently, the data terms used in the respective stages differ:
\begin{align}
E_\text{data}(c,\pose_t,\shape) = 
\begin{cases}
E_\text{detection}(c, \pose_t,\shape) & \mbox{for Stage I (\cref{sec:detections})}\\
E_\text{contour}(c, \pose_t,\shape) & \mbox{for Stage II (\cref{sec:contours})}\text{.}
\end{cases}
\label{egn:dataTerms}
\end{align}
The analytic form of all terms as well as the smoothness in all model parameters allows efficient optimization by gradient descent.
\NEW{In our experiments we apply the conditioned gradient descent method of Stoll et al.~\cite{stoll2011iccv}.}

\subsection{Stage I -- Initial estimation}
\label{sec:detections}
\label{sec:detectionFit}

We employ the discriminative ConvNet-based body-part detector by Tompson et al. \cite{tompson2014joint} to estimate the approximate 2D skeletal joint positions. 
The detector is independently applied to each input frame $\I_{c,t}$, and outputs heat maps of joint location probability $\detection_{c,t,j}$ for each joint $j$ in frame $t$ seen from camera $c$.
Importantly, the detector discriminates the joints on the left and right side of the body (see \cref{fig:overview}).
The detections exhibit noticeable spatial and temporal uncertainty, but are nonetheless a valuable cue for an initial space-time optimization solve.
The output heat maps are in general multi-modal due to detection ambiguities, but also in the presence of multiple people, e.g. in the background.

To infer the poses $\Poseset$ and an initial guess for the body shape of the subject, we optimize \cref{egn:objective} with data term $E_\text{detection}$. 
It measures the overlap of the heat maps $\detection$ with the projected skeleton joints.
Each joint in the model skeleton has an attached joint Gaussian (\cref{sec:bodyModel}), and the overlap with the corresponding heat map is maximized using the visibility model of Rhodin et al. \cite{rhodin2015iccv}.
We use a hierarchical approach by first optimizing the torso joints, followed by optimizing the limbs; please see the supplemental document for details.
The optimization is initialized with the average human shape $(\bar{\volume},\bar{\bones})$ in T-pose, at the center of the capture volume.
We assume a single person in the capture volume; people in the background are implicitly ignored, as they are typically not visible from all cameras and are dominated by the foreground actor.

Please note that bone lengths $\bones(\shape)$ and volume $\volume(\shape)$ are determined through $\shape$, hence, Stage~I yields a rough estimate of $\volume$.
In Stage~II, we use more informative image constraints than pure joint locations to better estimate volumetric extent. %

\subsection{Stage II -- Contour-based refinement}
\label{sec:contours}

The pose $\Poseset$ and shape $\shape$ found in the previous stage 
are now refined by using a new density-based contour model in the alignment energy.
This model explains the spatial image gradients formed at the edge of the projected model, between actor and background, and thus bypasses the need for silhouette extraction, which is difficult for general scenes. %
To this end, we extend the ray-casting image formation model of Rhodin et al.~\cite{rhodin2015iccv}, as summarized in the following paragraph, and subsequently explain how to use it in the contour data term $E_\text{contour}$.

\inlineheading{Ray-casting image formation model}
Each image pixel spawns a ray that starts at the camera center $\cam$ and points in direction $\n$.
The visibility of a particular model Gaussian $G_q$ along the ray $(\cam,\n)$ is defined as
\begin{align}
\V_q(\cam,\n) = \int_{0}^{\infty} \!\!   \exp\!\left(-\int_0^s \sum_i G_i(\cam+t \n) \dt \right) G_q(\cam+s\n) \ds
\text{.}
\label{egn:PreciseGaussianVisibility}
\end{align}
This equation models light transport in a heterogeneous translucent medium \cite{Cerezo2005}, i.e. $\V_q$ is the fraction of light along the ray that is absorbed by Gaussian $G_q$.
The original paper~\cite{rhodin2015iccv} describes an analytic approximation to \cref{egn:PreciseGaussianVisibility} by sampling the outer integral.

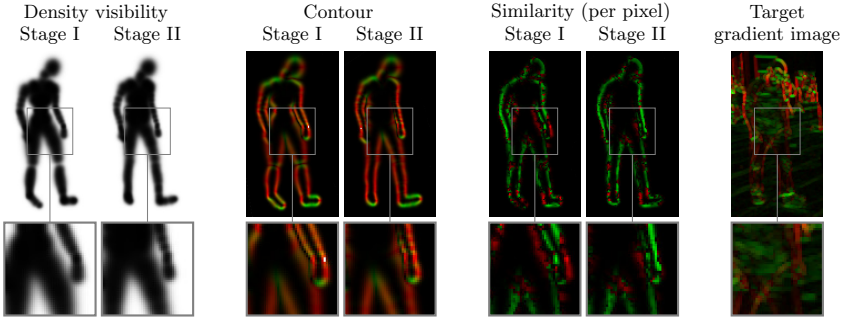
\begin{figure}[t!]
\centering%
		\begin{tikzpicture}[tight background,%
		image/.style={inner sep=0pt},
		smalltext/.style={scale=0.8},
		spy using outlines={rectangle, size=0.1\linewidth, magnification=2, connect spies},
		subcaption/.style={inner xsep=0.75mm, inner ysep=0.75mm, scale=0.6, above right},]
		\def\padding{2pt}
		\def\paddingB{20pt}
		\newcommand{\subfig}[2]{\includegraphics[width=0.1\linewidth,angle=0,origin=c,trim=110px 0px 250px 50px,clip]{images/Overview/#1#2.jpg}}

		\node[image,right=\padding] (B1) at (0,0) {\subfig{INIT_Silhouette_fblob0_display3/image_c0_f}{156}};
		\node[image,right=\padding] (B2) at (B1.east) {\subfig{REFINE_MODEL_Silhouette_fblob0_display3/image_c0_f}{156}};
		
		\node[image,right=\paddingB] (C1) at (B2.east) {\subfig{INIT_Silhouette_fblob0_display8/image_c0_f}{156}};
		\node[image,right=\padding] (C2) at (C1.east) {\subfig{REFINE_MODEL_Silhouette_fblob0_display8/image_c0_f}{156}};
		
		\node[image,right=\paddingB] (E1) at (C2.east) {\subfig{INIT_Silhouette_fblob0_display4/image_c0_f}{156}};
		\node[image,right=\padding] (E2) at (E1.east) {\subfig{REFINE_MODEL_Silhouette_fblob0_display4/image_c0_f}{156}};;
		
		\node[image,right=\paddingB] (T1) at (E2.east) {\subfig{INIT_Silhouette_fblob0_display7/image_c0_f}{156}};
		
		\node[smalltext, above=10pt] at (B1.north east) {Density visibility};
		\node[smalltext, above=10pt] at (C1.north east) {\phantom{g} Contour \phantom{g}};
		\node[smalltext, above=10pt] at (E1.north east) {Similarity (per pixel)};
		\node[smalltext, above=10pt] at (T1.north) {Target};
		
		\node[smalltext, above=0pt] at (B1.north) {Stage I};
		\node[smalltext, above=0pt] at (B2.north) {Stage II};
		\node[smalltext, above=0pt] at (C1.north) {Stage I};
		\node[smalltext, above=0pt] at (C2.north) {Stage II};
		\node[smalltext, above=0pt] at (E1.north) {Stage I};
		\node[smalltext, above=0pt] at (E2.north) {Stage II};
		\node[smalltext, above=0pt] at (T1.north) {gradient image};
		
		\spy[gray] on ($ (B1) + (0pt, 1pt) $) in node [below=2pt] at (B1.south);
		\spy[gray] on ($ (B2) + (0pt, 1pt) $) in node [below=2pt] at (B2.south);
		\spy[gray] on ($ (C1) + (0pt, 1pt) $) in node [below=2pt] at (C1.south);
		\spy[gray] on ($ (C2) + (0pt, 1pt) $) in node [below=2pt] at (C2.south);
		\spy[gray] on ($ (E1) + (0pt, 1pt) $) in node [below=2pt] at (E1.south);
		\spy[gray] on ($ (E2) + (0pt, 1pt) $) in node [below=2pt] at (E2.south);
		\spy[gray] on ($ (T1) + (0pt, 1pt) $) in node [below=2pt] at (T1.south);

		\end{tikzpicture}
	\caption{Contour refinement to image gradients through per-pixel similarity. Contour color indicates direction, green and red energy indicate agreement and disagreement between model and image gradients, respectively. Close-ups highlight the shape optimization: left arm and right leg pose are corrected in Stage II.}
	\label{fig:contourRefinement}
\end{figure}

Different to their work, we apply this ray casting model to infer the visibility of the background, $\B(\cam,\n) \!=\! 1 \!-\! \sum_q \V_q(\cam,\n)$.
Assuming that the background is infinitely distant, $\B$ is the fraction of light not absorbed by the Gaussian model: %
\begin{align}
\B(\cam,\n) = \exp\!\left(-\int_0^{\infty} \sum_q G_q(\cam+t \n) \dt \right)
= \exp\!\left(- \sqrt{2\pi} \sum_q \sR_q \cR_q  \right)
\text{.}
\label{egn:BackgroundVisibility}
\end{align}
This analytic form is obtained without sampling, but rather it stems from the Gaussian parametrization: the density along ray $(\cam,\n)$ though 3D Gaussian $G_q$ is a 1D Gaussian with standard deviation $\sR_q \!=\! \s_q$ and density maximum $\cR_q \!=\! c_q \cdot \exp\!\left(- \frac{(\m_q \!- \cam)^{\!\top} (\m_q \!- \cam) - ((\m_q \!- \cam)^{\!\top} \n)^2}{2 \s_q^2}\right)$, and the integral over the Gaussian density evaluates to a constant (when the negligible density behind the camera is ignored). %
A model visibility example is shown in \cref{fig:contourRefinement} left.

To extract the contour of our model, we compute the gradient of the background visibility $\BG \!=\! (\frac{\partial \B}{\partial u},\frac{\partial \B}{\partial v})^{\!\top}$ with respect to pixel location $(u,v)$:
\begin{align}
\BG  = \B \sqrt{2 \pi} \sum_q \frac{\cR_q}{\sR_q} (\m_q \!- \cam)^{\!\top}\n  (\m_q \!- \cam)^{\!\top} \nG
\text{.}
\end{align}
$\BG$ forms a 2D vector field, where the gradient direction points outwards from the model, and the magnitude forms a ridge at the model contour, see \cref{fig:contourRefinement} center.
In (calibrated pinhole) camera coordinates, the ray direction thus depends on the 2D pixel location $(u,v)$ by $\n \!=\! \frac{(u,v,1)^{\!\top}}{\norm{(u,v,1)}_2}$ and $\nG \!=\! (\frac{\partial \n}{\partial u}, \frac{\partial \n}{\partial v})^{\!\top}$.

In contrast to Rhodin et al.'s visibility model~\cite{rhodin2015iccv}, our model is specific to background visibility, but more accurate and efficient to evaluate.
It does not require sampling along the ray to obtain a smooth analytic form, has linear complexity in the number of model Gaussians instead of their quadratic complexity, and improves execution time by an order of magnitude.

\inlineheading{Contour energy}
\label{sec:contourSimilarity}
To refine the initial pose and shape estimates from Stage~I (\cref{sec:detections}), we optimize \cref{egn:objective} with a new contour data term $E_\text{contour}$, 
to estimate the per-pixel similarity of model and image gradients:
\begin{align}
E_\text{contour} (c, \pose_t, \shape) = \sum_{(u,v)} E_\text{sim}(c, \pose_t, \shape, u, v) + E_\text{flat}(c, \pose_t, \shape, u, v)\text{.}
\end{align}
In the following, we omit the arguments $(c, \pose_t, \shape, u, v)$ for better readability.
$E_\text{sim}$ measures the similarity between the gradient magnitude $\norm{\IG}_2$ in the input image and the contour magnitude $\norm{\BG}_2$ of our model, and penalizes orientation misalignment (contours can be in opposite directions in model and image):
\begin{align}
E_\text{sim} &= - \norm{\BG}_2 \norm{\IG}_2 \cos \!\big(2 \angle(\BG, \IG) \big)
\text{.}
\end{align}
The term $E_\text{flat}$ models contours forming in flat regions with gradient magnitude smaller than $\delta_\text{low} \!=\! 0.1$:
\begin{align}
E_\text{flat} &=
\norm{\BG}_2 \max(0,\delta_\text{low} - \norm{\IG}_2)
\text{.}
\label{egn:contourSimilarity}
\end{align}
We compute spatial image gradients $\IG=(\frac{\partial \I}{\partial u}, \frac{\partial \I}{\partial v})^\top$ using the Sobel operator, smoothed with a Gaussian ($\sigma \!=\! 1.1\,\text{px}$), summed over the RGB channels and clamped to a maximum of $\delta_\text{high}=0.2$.

\begin{figure}[t!]
\centering
		\begin{tikzpicture}[tight background,%
		image/.style={inner sep=0pt},
		subcaption/.style={inner xsep=0.75mm, inner ysep=0.75mm, scale=0.6, above right},]
		\def\padding{0pt}
		\newcommand{\subfigA}[2]{\includegraphics[height=0.14\linewidth,angle=0,origin=c,trim=190px 100px 400px 10px,clip]{images/VisualHullComparison/#1#2.jpg}}
		\newcommand{\subfigB}[2]{\includegraphics[height=0.14\linewidth,angle=0,origin=c,trim=260px 175px 430px 90px,clip]{images/VisualHullComparison/#1#2.jpg}}
		\newcommand{\subfigC}[2]{\includegraphics[height=0.14\linewidth,angle=0,origin=c,trim=420px 90px 250px 140px,clip]{images/VisualHullComparison/#1#2.jpg}}
		\newcommand{\subfigD}[2]{\includegraphics[height=0.14\linewidth,angle=0,origin=c,trim=550px 50px 10px 0px,clip]{images/VisualHullComparison/#1#2.jpg}}
		\newcommand{\subfigAA}[2]{\includegraphics[height=0.14\linewidth,angle=0,origin=c,trim=180px 110px 250px 10px,clip]{images/Volleyball/#1#2.jpg}}
		\newcommand{\subfigBB}[2]{\includegraphics[height=0.14\linewidth,angle=0,origin=c,trim=225px 50px 120px 10px,clip]{images/Volleyball/#1#2.jpg}}
		\newcommand{\subfigCC}[2]{\includegraphics[height=0.14\linewidth,angle=0,origin=c,trim=130px 50px 200px 20px,clip]{images/Volleyball/#1#2.jpg}}
		\newcommand{\subfigBBB}[2]{\includegraphics[height=0.14\linewidth,angle=0,origin=c,trim=250px 80px 120px 50px,clip]{images/Volleyball/#1#2.jpg}}
		\newcommand{\subfigCCC}[2]{\includegraphics[height=0.14\linewidth,angle=0,origin=c,trim=220px 80px 200px 65px,clip]{images/Volleyball/#1#2.jpg}}
		
		\newcommand{\subfigJA}[2]{\includegraphics[height=0.14\linewidth,angle=0,origin=c,trim=95px 50px 200px 5px,clip]{images/VisualHullComparison/#1#2.jpg}}
		\newcommand{\subfigJB}[2]{\includegraphics[height=0.14\linewidth,angle=0,origin=c,trim=130px 87.5px 215px 45px,clip]{images/VisualHullComparison/#1#2.jpg}}
		\newcommand{\subfigJC}[2]{\includegraphics[height=0.14\linewidth,angle=0,origin=c,trim=210px 45px 125px 70px,clip]{images/VisualHullComparison/#1#2.jpg}}
		\newcommand{\subfigJD}[2]{\includegraphics[height=0.14\linewidth,angle=0,origin=c,trim=275px 25px 5px 0px,clip]{images/VisualHullComparison/#1#2.jpg}}
		\newcommand{\subfigJAA}[2]{\includegraphics[height=0.14\linewidth,angle=0,origin=c,trim=90px 55px 125px 5px,clip]{images/Volleyball/#1#2.jpg}}
		\newcommand{\subfigJBB}[2]{\includegraphics[height=0.14\linewidth,angle=0,origin=c,trim=112.5px 25px 60px 5px,clip]{images/Volleyball/#1#2.jpg}}
		\newcommand{\subfigJCC}[2]{\includegraphics[height=0.14\linewidth,angle=0,origin=c,trim=65px 25px 100px 10px,clip]{images/Volleyball/#1#2.jpg}}
		\newcommand{\subfigJBBB}[2]{\includegraphics[height=0.14\linewidth,angle=0,origin=c,trim=125px 40px 60px 25px,clip]{images/Volleyball/#1#2.jpg}}
		\newcommand{\subfigJCCC}[2]{\includegraphics[height=0.14\linewidth,angle=0,origin=c,trim=110px 40px 100px 32.5px,clip]{images/Volleyball/#1#2.jpg}}

		\node[image,right=\padding] (Af0) at (0,0) {\subfigJA{Input/image_c1_f}{28}};
		\node[image,right=\padding] (Bf0) at (Af0.east) {\subfigJB{Input/image_c4_f}{28}};
		\node[image,right=\padding] (Cf0) at (Bf0.east) {\subfigJC{Input/image_c5_f}{28}};
		\node[image,right=\padding] (Df0) at (Cf0.east) {\subfigJD{Input/image_c7_f}{28}};

		\node[image,right=16pt] (Af1) at (Df0.east) {\subfigA{Density/image_c1_f}{28}};
		\node[image,right=\padding] (Bf1) at (Af1.east) {\subfigB{Density/image_c4_f}{28}};
		\node[image,right=\padding] (Cf1) at (Bf1.east) {\subfigC{Density/image_c5_f}{28}};
		\node[image,right=\padding] (Df1) at (Cf1.east) {\subfigD{Density/image_c7_f}{28}};

		\node[image,right=16pt] (Af2) at (Df1.east) {\subfigJA{Input/image_composition_c1_f}{28}};
		\node[image,right=\padding] (Bf2) at (Af2.east) {\subfigJB{Input/image_composition_c4_f}{28}};
		\node[image,right=\padding] (Cf2) at (Bf2.east) {\subfigJC{Input/image_composition_c5_f}{28}};
		\node[image,right=\padding] (Df4) at (Cf2.east) {\subfigJD{Input/image_composition_c7_f}{28}};

		\node[above=5pt,anchor=west] at (Af0.north west) {\scriptsize Input views};
		\node[above=5pt,anchor=west] at (Af1.north west) {\scriptsize Colored density};
		\node[above=5pt,anchor=west] at (Af2.north west) {\scriptsize Actor skeleton and mesh};

		\node[image,below=2pt,anchor=north west] (f0) at (Af0.south west) {\subfigJAA{Input/image_c0_f}{334}};
		\node[image,right=\padding] (Bf0) at (f0.east) {\subfigJBB{Input/image_c3_f}{334}};
		\node[image,right=\padding] (Cf0) at (Bf0.east) {\subfigJCC{Input/image_c4_f}{334}};

		\node[image,below=2pt,anchor=north west] (f1) at (Af1.south west) {\subfigAA{Density/image_c0_f}{334}};
		\node[image,right=\padding] (Bf1) at (f1.east) {\subfigBBB{Density/image_c3_f}{334}};
		\node[image,right=\padding] (Cf1) at (Bf1.east) {\subfigCCC{Density/image_c4_f}{334}};

		\node[image,below=2pt,anchor=north west] (f2) at (Af2.south west) {\subfigJAA{Input/image_composition_c0_f}{334}};
		\node[image,right=\padding] (Bf2) at (f2.east) {\subfigJBBB{Input/image_composition_c3_f}{334}};
		\node[image,right=\padding] (Cf2) at (Bf2.east) {\subfigJCCC{Input/image_composition_c4_f}{334}};

		\end{tikzpicture}
	\caption{Reconstruction of challenging outdoor sequences with complex motions from only 3–4 views, showing accurate shape and pose reconstruction.%
	}
	\label{fig:outdoor}
\end{figure}
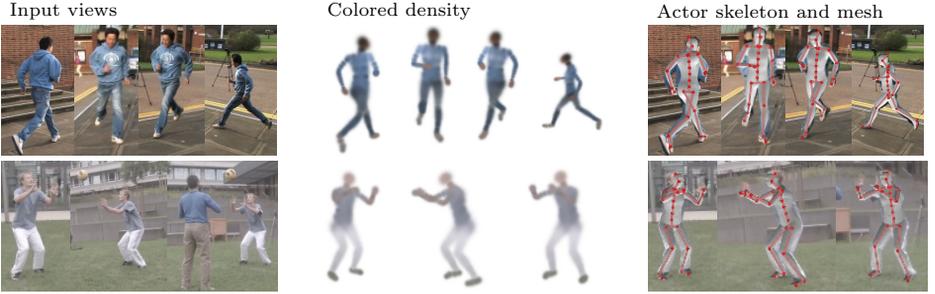

\inlineheading{Appearance estimation}
\label{sec:appearance}
Our method is versatile: given the shape and pose estimates from Stage~II, we can also estimate a color for each Gaussian.
This is needed by earlier tracking methods that use similar volume models, but color appearance-based alignment energies \cite{rhodin2015iccv,stoll2011iccv} – we compare against them in our experiments.
Direct back-projection of the image color onto the model suffers from occasional reconstruction errors in Stages I and II.
Instead, we compute the weighted mean color $\bar{a}_{q,c}$ over all pixels separately for each Gaussian $G_q$ and view $c$, where the contribution of each pixel is weighted by the Gaussian's visibility $\V_q$ (\cref{egn:PreciseGaussianVisibility}).
Colors $\bar{a}_{q,c}$ are taken as candidates from which outliers are removed by iteratively computing the mean color and removing the largest outlier (in Euclidean distance).
In our experiments, removing 50\% of the candidates leads to consistently clean color estimates, as shown in \cref{fig:outdoor,fig:extremeBodyShapes}.

\IGNORE{
As our scene and image formation models are entirely volumetric, %
\CR{I wouldn't say the image formation model is `volumetric'}
surfaces only implicitly appear at isosurfaces of the density.
To enable application where explicit meshes are required, like for further refinement or animation, a surface could be reconstructed using marching cubes on the density.
Instead, we propose a direct coupling, where the reference mesh is deformed by the underlying Gaussian representation.
Given an unseen new Gaussian actor volume $\volume_j$ (different from any mesh database instance), we take the affine transformation of each Gaussian from reference $\volume_\text{ref}$ to $\volume_j$, and apply the same to the reference mesh.
Similar to linear blend skinning \cite{Lewis:2000} \CR{check ref}, the transformations are applied in a weighted manner to all neighboring vertices, weighted by the neighborhood weights inferred in \cref{sec:bodyModel}.
This interpretation of skinning builds a linear deformation model parametrized by the volume  $\volume$.
On our database, it showed outstanding generalization properties for shape extrapolation, of equal quality to a deformation gradient representation (like SCAPE \cite{Anguelov2005}), which would, however, require solving a Poisson System, and with fewer smoothness artifacts but equal in performance than using vertex positions directly, see supplemental video \HR{or Fig.}.
}

\section{Evaluation}
\label{sec:eval}
We evaluate our method on 11 sequences of publicly available datasets with large variety, both indoor and outdoor, and show comparisons to state-of-the-art methods  (see supplementary document for dataset details). 
The quality of pose and shape reconstruction is best assessed in the supplemental video, 
where we also apply and compare our reconstructions to tracking with the volumetric Gaussian representations of Rhodin et al. \cite{rhodin2015iccv} and Stoll et al. \cite{stoll2011iccv}.

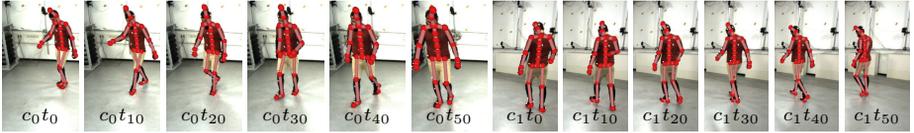
\begin{figure}[t]
	
	\begin{tikzpicture}[tight background,%
	image/.style={inner sep=0pt},
	subcaption/.style={inner xsep=0.75mm, inner ysep=0.75mm, scale=0.6, above right},]
	\def\padding{2pt}
	\newcommand{\subfigM}[2]{\includegraphics[height=0.14\linewidth,angle=0,origin=c,trim=25px 200px 320px 30px,clip]{images/ms-walk-twoViews/#1#2.jpg}}
	\newcommand{\subfigMM}[2]{\includegraphics[height=0.14\linewidth,angle=0,origin=c,trim=320px 200px 50px 30px,clip]{images/ms-walk-twoViews/#1#2.jpg}}
	
	\node[image,below=2mm] (Af0) at (0,0) {\subfigM{image_c5_f}{850}};
	\node[image,right=\padding] (Af1) at (Af0.east) {\subfigM{image_c5_f}{860}};
	\node[image,right=\padding] (Af2) at (Af1.east) {\subfigM{image_c5_f}{870}};
	\node[image,right=\padding] (Af3) at (Af2.east) {\subfigM{image_c5_f}{880}};
	\node[image,right=\padding] (Af4) at (Af3.east) {\subfigM{image_c5_f}{890}};
	\node[image,right=\padding] (Af5) at (Af4.east) {\subfigM{image_c5_f}{900}};
	
	\node[image,right=2pt] (Af6) at (Af5.east) {\subfigMM{image_c1_f}{850}};
	\node[image,right=\padding] (Af7) at (Af6.east) {\subfigMM{image_c1_f}{860}};
	\node[image,right=\padding] (Af8) at (Af7.east) {\subfigMM{image_c1_f}{870}};
	\node[image,right=\padding] (Af9) at (Af8.east) {\subfigMM{image_c1_f}{880}};
	\node[image,right=\padding] (Af10) at (Af9.east) {\subfigMM{image_c1_f}{890}};
	\node[image,right=\padding] (Af11) at (Af10.east) {\subfigMM{image_c1_f}{900}};
	
	\node[above=0pt] at (Af0.south) {\scriptsize $c_0 t_0$};
	\node[above=0pt] at (Af1.south) {\scriptsize $c_0 t_{10}$};
	\node[above=0pt] at (Af2.south) {\scriptsize $c_0 t_{20}$};
	\node[above=0pt] at (Af3.south) {\scriptsize $c_0 t_{30}$};
	\node[above=0pt] at (Af4.south) {\scriptsize $c_0 t_{40}$};
	\node[above=0pt] at (Af5.south) {\scriptsize $c_0 t_{50}$};
	
	\node[above=0pt] at (Af6.south) {\scriptsize $c_1 t_{0}$};
	\node[above=0pt] at (Af7.south) {\scriptsize $c_1 t_{10}$};
	\node[above=0pt] at (Af8.south) {\scriptsize $c_1 t_{20}$};
	\node[above=0pt] at (Af9.south) {\scriptsize $c_1 t_{30}$};
	\node[above=0pt] at (Af10.south) {\scriptsize $c_1 t_{40}$};
	\node[above=0pt] at (Af11.south) {\scriptsize $c_1 t_{50}$};
	\end{tikzpicture}
	\caption{We obtained accurate results even using only two views on the \texttt{Marker} sequence.}
	\label{fig:twocameras}
\end{figure}

\inlineheading{Robustness in general scenes}
We validate the robustness of our method on three outdoor sequences. 
On the \texttt{Walk} dataset \cite{elhayek2015efficient}, people move in the background, and background and foreground color are very similar. Our method is nevertheless able to accurately estimate shape and pose across 100 frames from 6 views, see \cref{fig:overview}.
\NEW{
We also qualitatively compare against the recent the model-free method of Mustafa \etal~\cite{Mustafa2015}. On the \texttt{Cathedral} sequence of Kim \etal~\cite{kim2014influence},} they achieve rough surface reconstruction using 8 cameras without the explicit need for silhouettes; in contrast, 4 views and 20 frames are sufficient for us to reconstruct shape and pose of a quick outdoor run%
, see \cref{fig:outdoor} (top) and supplementary material. 
Furthermore, we demonstrate reconstruction \NEW{of complex motions on} \texttt{Subject3} during a two-person volleyball play from only 3 views and 100 frames, see \cref{fig:outdoor} (bottom).
The second player was segmented out during Stage I, but Stage II was executed automatically.
Fully automatic model and pose estimation are even possible from only two views as we demonstrate on the \texttt{Marker} sequence \cite{elhayek2015efficient}, see \cref{fig:twocameras}.

\begin{figure}[t]
	\begin{tikzpicture}
	\draw (-5, 0) node[inner sep=0] {\includegraphics[width=4cm]{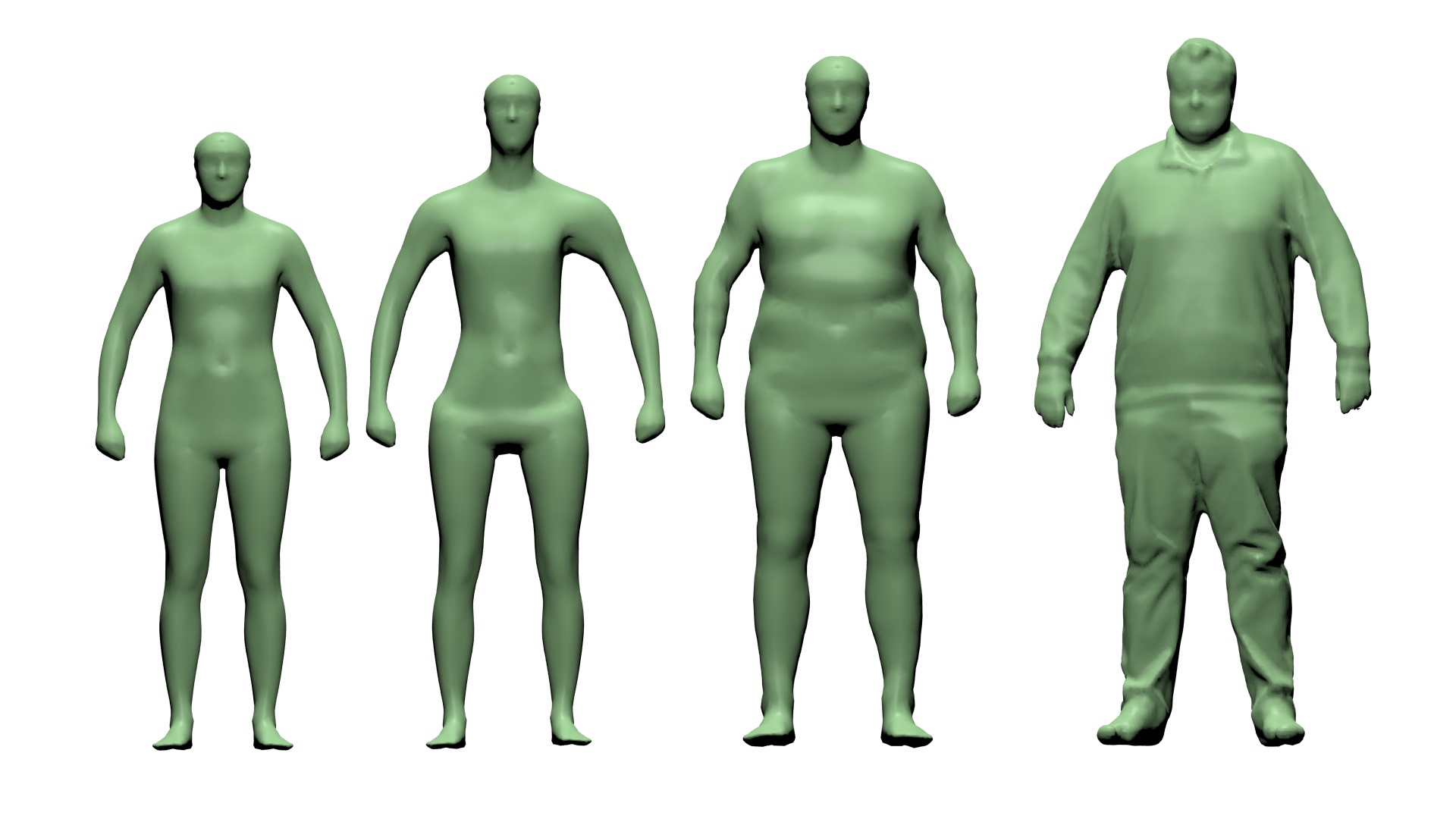}};
	\draw (-5, -1.3) node { \scriptsize{\textbf{(a)} \texttt{Subject1}}};
	\draw (-0.8, 0) node[inner sep=0] {\includegraphics[width=4cm]{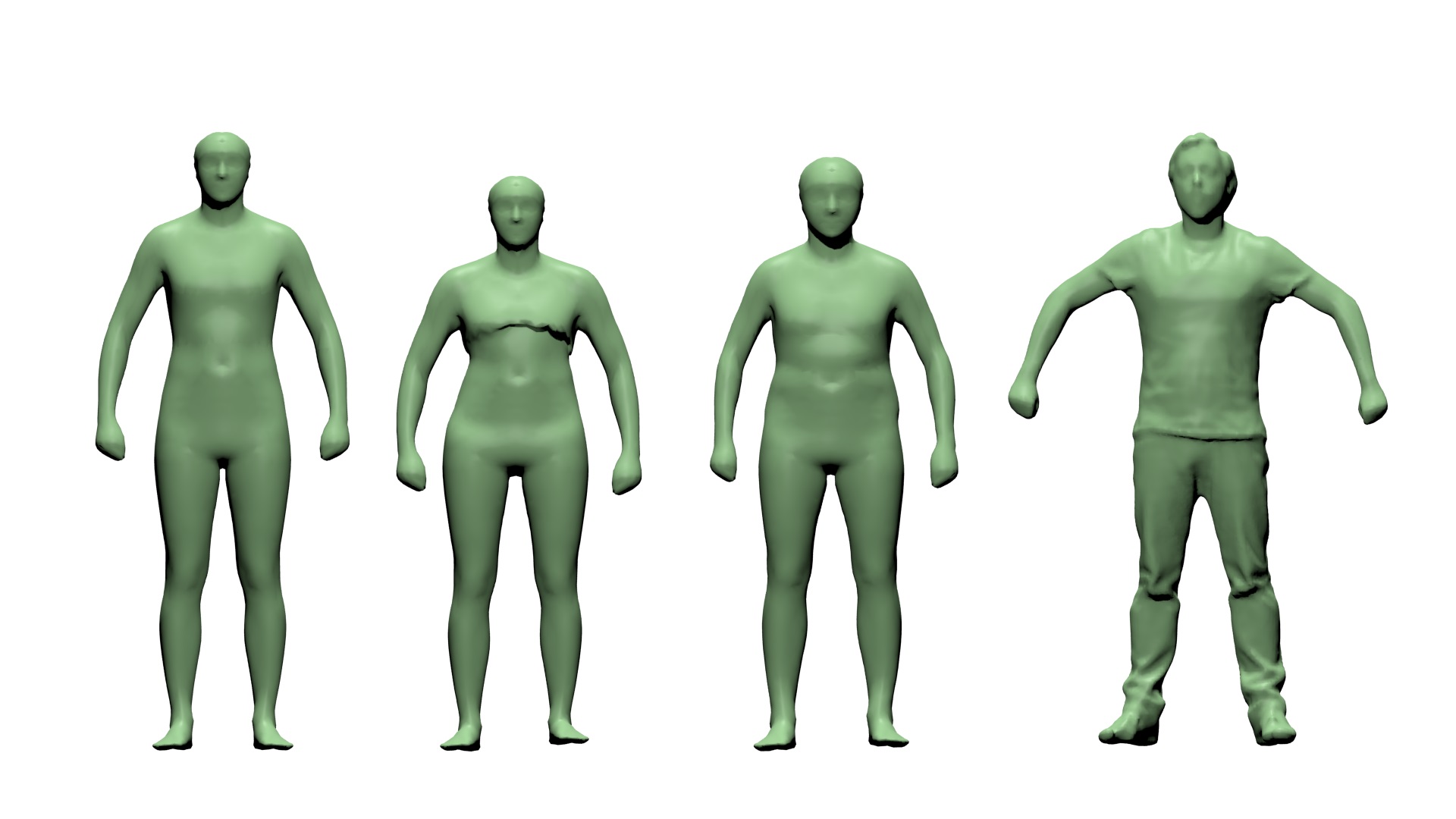}};
	\draw (-0.8, -1.3) node {\scriptsize{\textbf{(b)} \texttt{Subject2}}};
	\draw (3.5, 0) node[inner sep=0] {\includegraphics[width=4cm]{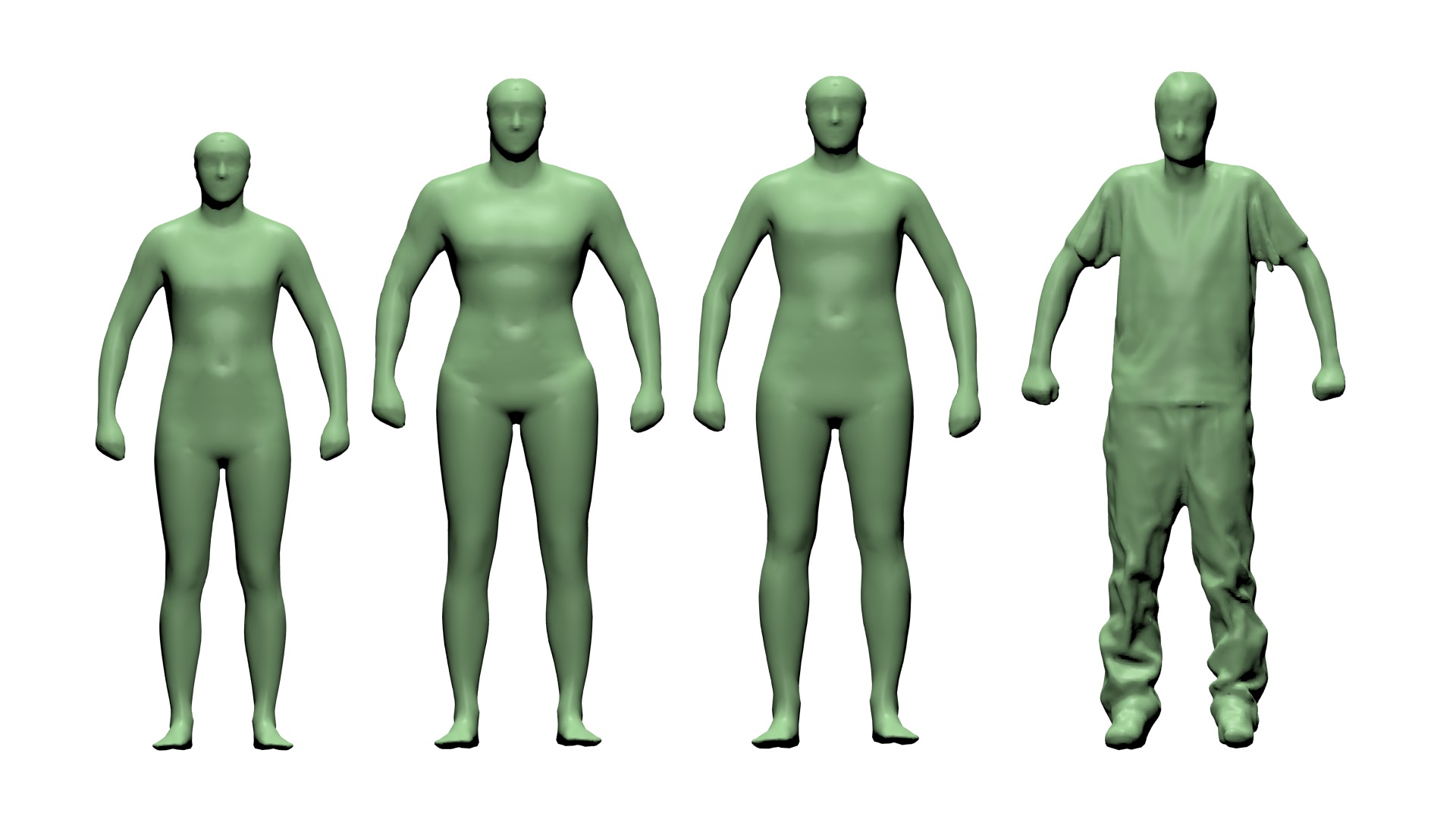}};
	\draw (3.5, -1.3) node {\scriptsize{\textbf{(c)} \texttt{Subject3}}};
	\end{tikzpicture}
	\caption{Visual comparison of estimated body shapes at the different stages. In each subfigure (from left to right): mean PCA $(\bar{\volume},\bar{\bones})$, Stage I, Stage II and ground-truth shape, respectively.}
	\label{fig:qualitative_shape_eval}
\end{figure}
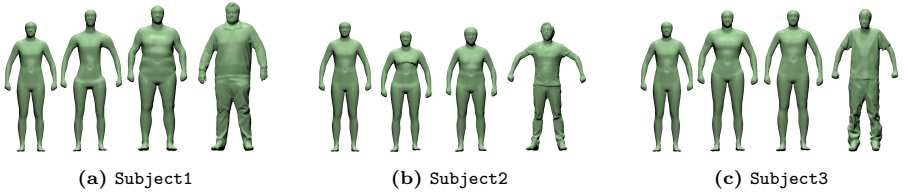

\begin{figure}[t]
	\begin{floatrow}
		\ttabbox[1\textwidth]{
			\resizebox{1\columnwidth}{!}{
				\setlength\extrarowheight{3pt}
				\begin{tabular}{| c ? C{0.9cm} | c | c | C{0.9cm} ? C{0.9cm} | c | c | C{0.9cm} ? C{0.9cm} | c | c | C{0.9cm} ? C{0.9cm} | c | c |  C{0.9cm} ? c }
					\cline{2-17}
					\multicolumn{1}{c?}{}& \multicolumn{4}{c?}{\textbf{\textcolor{green!70!black}{Chest size}} [cm]} & \multicolumn{4}{c?}{\textbf{\textcolor{blue!70!black}{Waist size}} [cm]} & \multicolumn{4}{c?}{\textbf{\textcolor{red!70!black}{Hip size}} [cm]} &  \multicolumn{4}{c?}{\textbf{Height} [cm]} & \multirow{7}{*}{\hspace{0.2cm}\includegraphics[width=0.12\textwidth]{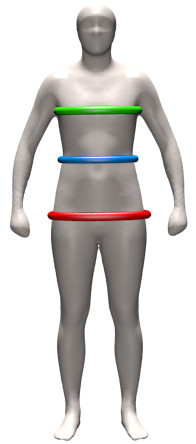}}\\
					\cline{2-17} 
					\multicolumn{1}{c?}{}& 	\cite{guan2009estimating} & Stage I & Stage II & GT & 	\cite{guan2009estimating}& Stage I & Stage II & GT& 	\cite{guan2009estimating} & Stage I & Stage II & GT & \cite{guan2009estimating} & Stage I & Stage II & ~GT~\\
					\Cline{1.1pt}{1-17}
					\texttt{Pose1} & 92.7 & -- & 92.8 & 92.6 & 79.6 & -- & 82.5 & 80.2 & --& -- & 98.0 & -- &-- & -- & 183.2 & 185.0\\
					\Cline{1pt}{1-17}
					\texttt{Pose2} & 87.4 & -- & 91.3 & 91.6 & 78.5 & -- & 82.1 & 79.4 & --& -- & 98.9 & -- & --& -- & 181.9 & 185.0\\
					\Cline{1pt}{1-17}
					\texttt{Pose3} & 91.9 & -- & 93.5 & 91.4 & 76.9 & -- & 83.2 & 80.3 & --& -- & 101.3 & -- & -- & -- & 182.9 & 185.0\\
					\Cline{1pt}{1-17}
					\texttt{Subject1} & -- & 92.7 & 132.6 & 131.3 & -- & 76.7 & 127.1 & 132.7 & --& 108.2 & 135.4 & 136.1 &-- & 187.5 & 194.2 & 195.0\\
					\Cline{1pt}{1-17}
					\texttt{Subject2} & -- & 92.3 & 99.3 & 100.1 & -- & 77.3 & 90.6 & 96.5  & --&92.5 & 102.3 & 99.7 & -- & 168.5 & 162.5 & 162.0\\
					\Cline{1pt}{1-17}
				\end{tabular}
			}
		}
		{\caption{\label{tab:chest_waist_hips}%
				Quantitative evaluation of estimated shapes in different stages and comparison to Guan et al.'s results \cite{guan2009estimating}.
				We use three body measures (chest, waist and hips, as shown on the right) to evaluate predicted body shapes against the ground truth (GT) captured using a laser scan.}\label{table:body_measure}}
\end{floatrow}
\end{figure}

\inlineheading{Shape estimation accuracy}
To assess the accuracy of the estimated actor models, we tested our method on a variety of subjects performing general motions such as walking, kicking and gymnastics. 
Evaluation of estimated shape is performed in two ways: (1) the estimated body shape is compared against ground-truth measurements, and (2) the 3D mesh derived from Stage II is projected from the captured camera viewpoints to compute the overlap with a manually segmented foreground.
We introduce two datasets \texttt{Subject1} and \texttt{Subject2}, in addition to \texttt{Subject3}, with pronounced body proportions and ground-truth laser scans for quantitative evaluation.
\NEW{Please note that shape estimates are constant across the sequence and can be evaluated at sparse frames, while pose varies and is separately evaluated per frame in the sequel.}

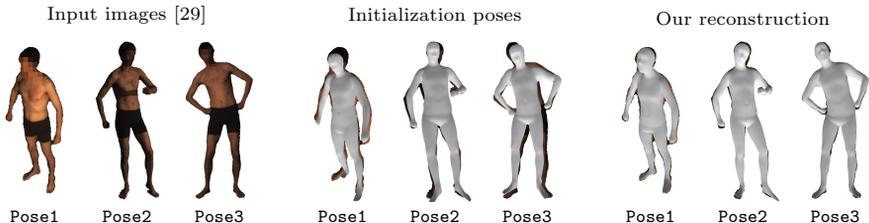
\begin{figure}[t!]
\centering
		\hspace{-0.5cm}
		\begin{tikzpicture}[tight background,%
		image/.style={inner sep=0pt},
		subcaption/.style={inner xsep=0.75mm, inner ysep=0.75mm, scale=0.6, above right},]
		\def\padding{0pt}
		\newcommand{\subfig}[2]{\includegraphics[width=0.1\linewidth,angle=0,origin=c,trim=80px 0px 80px 0px,clip]{images/monocularReconstruction/peng2009/#1#2.jpg}}
		
		\node(image_c0_f-1) at (0,0) {};
		\foreach \img/\prev in {0/-1,1/0,2/1}
		{
			\node[image,right=\padding] (image_c0_f\img) at (image_c0_f\prev.east) {\subfig{image_c0_f}{\img}};
		}
		\node[above=2pt] at (image_c0_f1.north) {\scriptsize Input images \cite{guan2009estimating}};
		\node[below=0pt] at (image_c0_f0.south) {\scriptsize \texttt{Pose1}};
		\node[below=0pt] at (image_c0_f1.south) {\scriptsize \texttt{Pose2}};
		\node[below=0pt] at (image_c0_f2.south) {\scriptsize \texttt{Pose3}};
		
		\node(image_c0_f-1) at ($(image_c0_f2.east)+(3mm,0)$) {};
		\foreach \img/\prev in {0/-1,1/0,2/1}
		{
			\node[image,right=\padding] (image_c0_f\img) at (image_c0_f\prev.east) {\subfig{InputPoses/image_composition_c0_f}{\img}};
		}
		\node[above=2pt] at (image_c0_f1.north) {\scriptsize Initialization poses};
		\node[below=0pt] at (image_c0_f0.south) {\scriptsize \texttt{Pose1}};
		\node[below=0pt] at (image_c0_f1.south) {\scriptsize \texttt{Pose2}};
		\node[below=0pt] at (image_c0_f2.south) {\scriptsize \texttt{Pose3}};
		
		\node(image_c0_f-1) at ($(image_c0_f2.east)+(3mm,0)$) {};
		\foreach \img/\prev in {0/-1,1/0,2/1}
		{
			\node[image,right=\padding] (image_c0_f\img) at (image_c0_f\prev.east) {\subfig{PerFrameOptimization/image_composition_c0_f}{\img}};
		}
		\node[above=2pt] at (image_c0_f1.north) {\scriptsize Our reconstruction};
		\node[below=0pt] at (image_c0_f0.south) {\scriptsize \texttt{Pose1}};
		\node[below=0pt] at (image_c0_f1.south) {\scriptsize \texttt{Pose2}};
		\node[below=0pt] at (image_c0_f2.south) {\scriptsize \texttt{Pose3}};
		\end{tikzpicture}
	\caption{%
		Monocular reconstruction experiment.
		Our reconstruction (right) shows high-quality contour alignment, and improved pose and shape estimates.}
	\label{fig:monocularReconstruction}
\end{figure}

The shape accuracy is evaluated by measurements of chest, waist and hip circumference.
\texttt{Subject1} and \texttt{Subject2} are captured indoor and are processed using 6 cameras and 40 frames \IGNORE{equally}\NEW{uniformly} sampled over 200 frames.
\texttt{Subject3} is an outdoor sequence and only 3 camera views are used, see \cref{fig:outdoor}.
All subjects are reconstructed with high quality in shape, skeleton dimensions and color, despite inaccurately estimated poses in Stage I for some frames.
We only observed little variation dependent on the performed motions, \ie a simple walking motion is sufficient, but bone length estimation degrades if joints are not sufficiently articulated during performance.
All estimates are presented quantitatively in \cref{tab:chest_waist_hips} and qualitatively in \cref{fig:qualitative_shape_eval}.
In addition, we compare against Guan \etal~\cite{guan2009estimating} on their single-camera and single-frame datasets \texttt{Pose1}, \texttt{Pose2} and \texttt{Pose3}.
Stage~I requires multi-view input and was not used; instead, we manually initialized the pose roughly, as shown in \cref{fig:monocularReconstruction}, and body height is normalized to 185\,cm \cite{guan2009estimating}.
Our reconstructions are within the same error range, demonstrating that Stage~II is well suited even for monocular shape and pose refinement. 
Our reconstruction is accurate overall, with a mean error of only $2.3\pm1.9$\,cm, measured across all sequences with known ground truth.

\begin{figure}[t]
	\begin{floatrow}
		\TopFloatBoxes
		\ffigbox[0.72\textwidth]{%
			\begin{tikzpicture}[image/.style={inner sep=0pt},]
				\node[image,right=0pt] (F0) at (0,0)		
				{\includegraphics[width=0.065\textwidth,trim=160px 40px 160px 40px,clip]{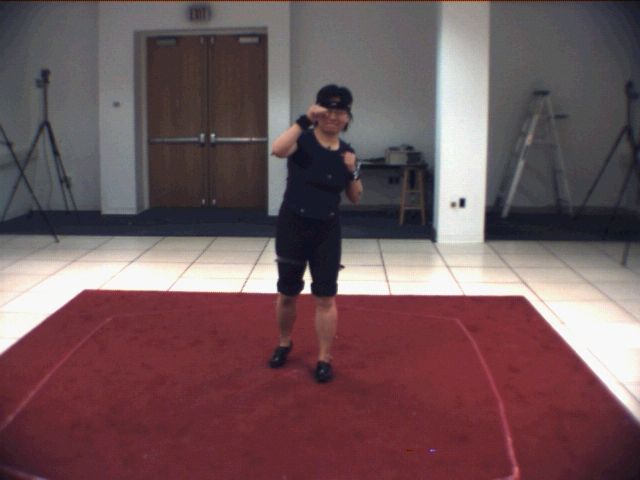}};
				\node[scale=0.7,below=2pt] at (F0.south) {Input};
				\node[image,right=2pt] (F1) at (F0.east)		
				{\includegraphics[width=0.065\textwidth,trim=160px 40px 160px 40px,clip]{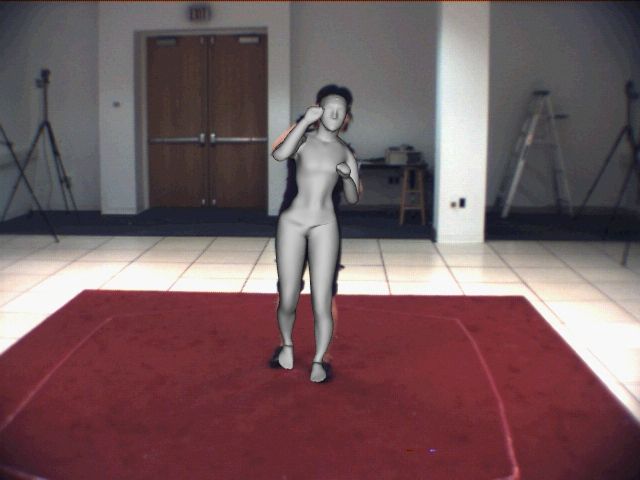}};
				\node[scale=0.7,below=9pt,right=2pt] at (F1.south) {Stage I};
				\node[image,right=0pt] (F2) at (F1.east)		
				{\includegraphics[width=0.065\textwidth,trim=160px 40px 160px 40px,clip]{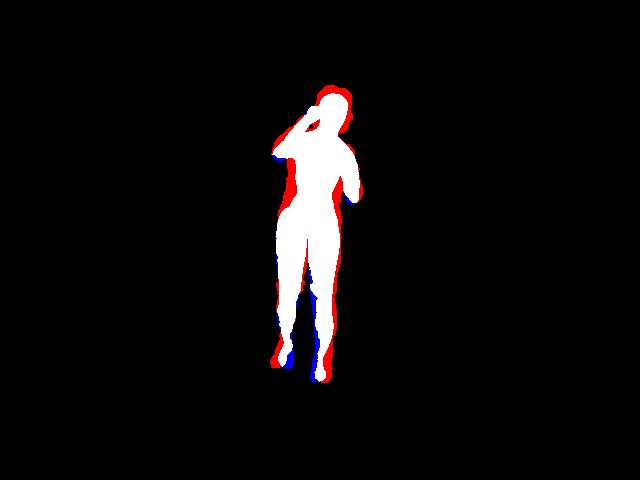}};
				\node[image,right=2pt] (F3) at (F2.east)
				{\includegraphics[width=0.065\textwidth,trim=160px 40px 160px 40px,clip]{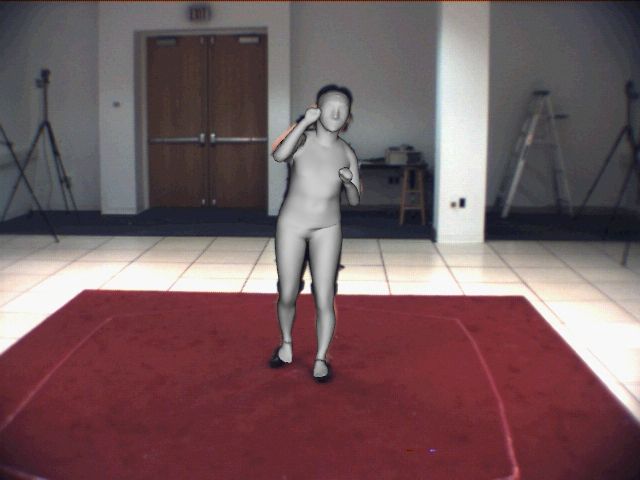}};
				\node[scale=0.7,below=9pt, right=1pt] at (F3.south) {Stage II};
				\node[image,right=0pt] (F4) at (F3.east)		
				{\includegraphics[width=0.065\textwidth,trim=160px 40px 160px 40px,clip]{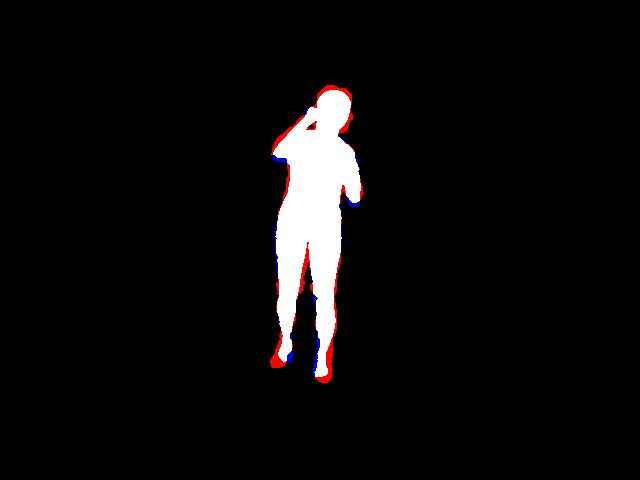}};

				\node[image,right=6pt] (F5) at (F4.east)
				{\includegraphics[width=0.065\textwidth,trim=10px 20px 310px 60px,clip]{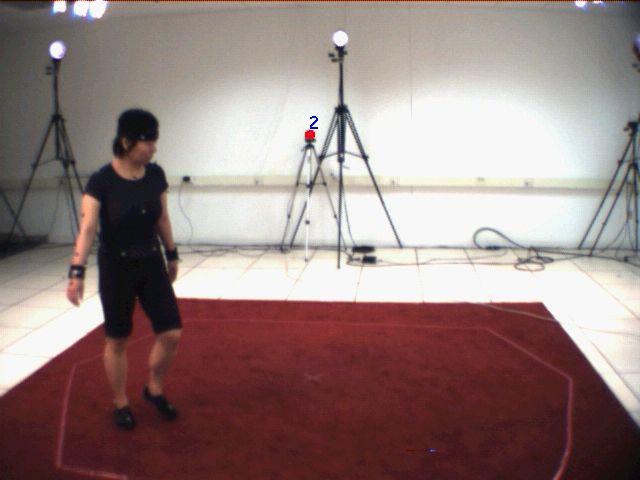}};
				\node[scale=0.7,below=2pt] at (F5.south) {Input};
				\node[image,right=2pt] (F6) at (F5.east)
				{\includegraphics[width=0.065\textwidth,trim=10px 20px 310px 60px,clip]{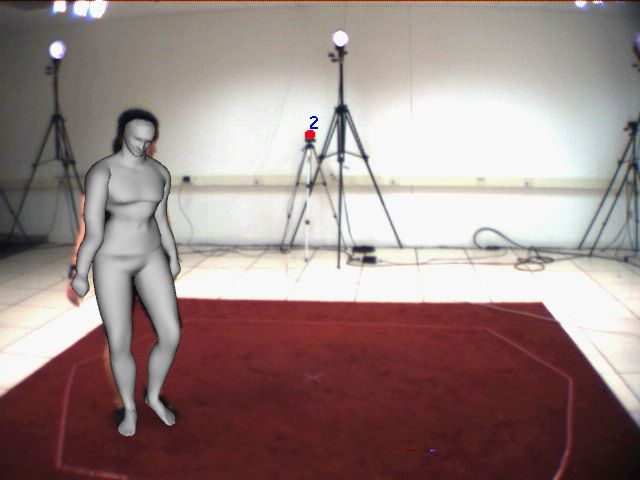}};
				\node[scale=0.7,below=9pt,right=2pt] at (F6.south) {Stage I};
				\node[image,right=0pt] (F7) at (F6.east)
				{\includegraphics[width=0.065\textwidth,trim=10px 20px 310px 60px,clip]{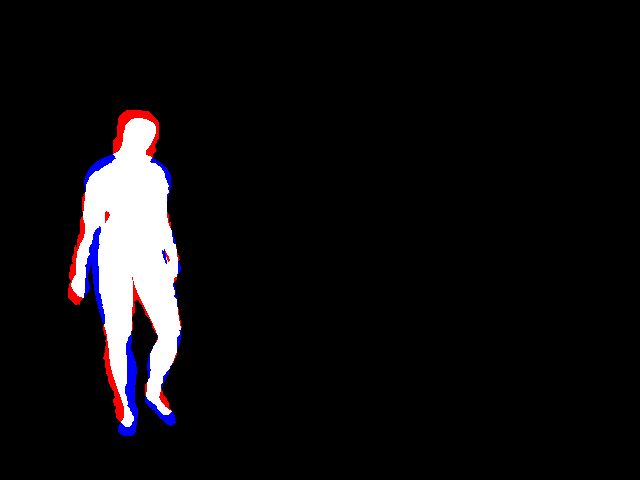}};
				\node[image,right=2pt] (F8) at (F7.east)
				{\includegraphics[width=0.065\textwidth,trim=10px 20px 310px 60px,clip]{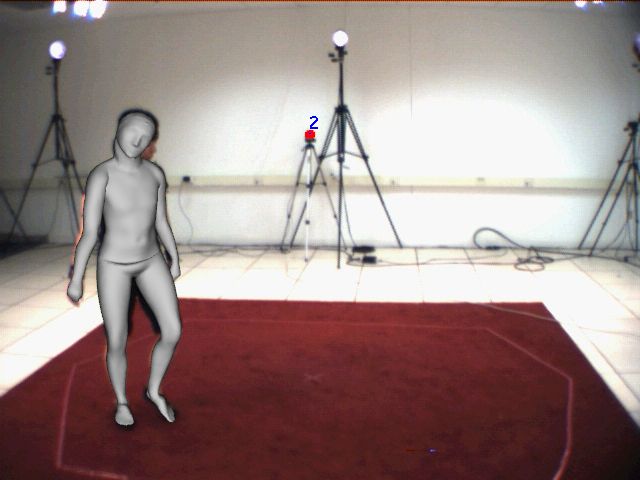}};
				\node[scale=0.7,below=9pt,right=1pt] at (F8.south) {Stage II};
				\node[image,right=0pt] (F9) at (F8.east)
				{\includegraphics[width=0.065\textwidth,trim=10px 20px 310px 60px,clip]{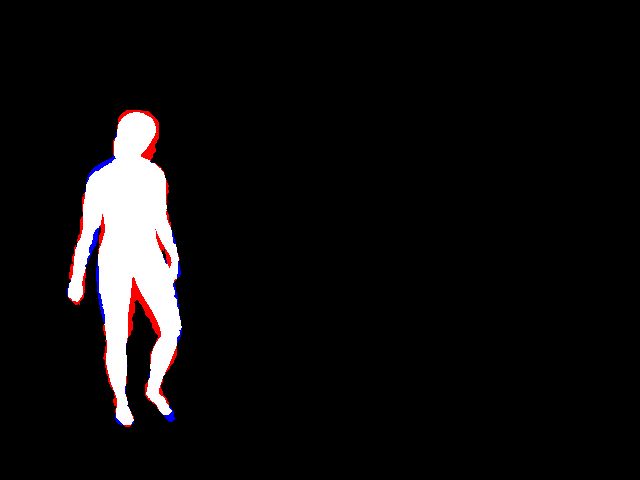}};
		\end{tikzpicture}%
		}
		{%
		\caption{Overlap of the estimated shape in Stages I and II for an input frame of \texttt{Box} (left) and \texttt{Walk} (right) sequences~\cite{sigal2010humaneva}. Note how the white area (correct estimated shape) significantly increases between Stage I and II, while blue (overestimation) and red (underestimation) areas decrease.}%
		\label{fig:operlap_qulitatively}
	}%
	\ttabbox[0.235\textwidth]{%
		\raisebox{0.8 cm}{%
			\resizebox{0.24\textwidth}{!}{%
				\setlength\extrarowheight{3pt}
				\begin{tabular}{|l |C{1.5cm} |C{1.5cm} |}
					\cline{2-3}
					\multicolumn{1}{c|}{} & \textbf{Precision} & \textbf{Recall} \\ 		
					\hline
					\texttt{Walk} Stage I & 87.43\% & 87.25\%  \\
					\hline
					\texttt{Walk} Stage II &95.18\% & 86.89\% \\
					\hline
					\texttt{Box} Stage I & 93.26\% & 81.11\% \\
					\hline
					\texttt{Box} Stage II & 95.42\% & 85.28\% \\
					\hline
				\end{tabular}
			}
		}
	}
	{\caption{\label{tab:overlap}%
			Quantitative evaluation of \cref{fig:operlap_qulitatively}.
			See \cref{sec:eval} for definitions of \textit{Precision} and \textit{Recall}.}}
\end{floatrow}
\end{figure}

On top of these sparse measurements (chest, waist and hips), we also evaluate silhouette overlap for sequences \texttt{Walk} and \texttt{Box} of subject 1 of the publicly available \texttt{HumanEva-I} dataset \cite{sigal2010humaneva}, using only 3 cameras.
We compute how much the predicted body shape overlaps the actual foreground (\textit{precision}) and how much of the foreground is overlapped by the model (\textit{recall}).
Despite the low number of cameras, low-quality images, and without requiring background subtraction, our reconstructions are accurate with 95\% precision and 85\% recall, and improve slightly on the results of B{\u{a}}lan et al. \cite{Balan:2007}.
Results
are presented in \cref{fig:operlap_qulitatively} and \cref{tab:overlap}.
Note that Stage~II significantly improves shape estimation.
\NEW{The temporal consistency and benefit of the model components are shown in the supplemental video on multiple frames evenly spread along the \textit{HumanEva-I} \texttt{Box} sequence.}

\inlineheading{Pose estimation accuracy}
Pose estimation accuracy is quantitatively evaluated on the public \texttt{HumanEva-I} dataset, where ground-truth data is available, see Table \ref{table:humaneva}.
We tested the method on the designated validation sequences \texttt{Walk} and \texttt{Box} of subject \texttt{S1}. 
Reconstruction quality is measured as the average Euclidean distance of estimated and ground-truth joint locations, frames with ground truth inaccuracies are excluded by the provided scripts.

Our pose estimation results are on par with state-of-the-art methods with 6–7\,cm average accuracy \cite{sigal2012loose,amin2013multi,belagiannis20143d,elhayek2015efficient}.
In particular, we obtain comparable results to Elhayek et al. \cite{elhayek2015efficient}, which however requires a separately initialized actor model.
Please note that Amin et al. \cite{amin2013multi} specifically trained their model on manually annotated sequences of the same subject in the same room. %
For best tracking performance, the ideal joint placement and bone lengths of the virtual skeleton may deviate from the real human anatomy, and may generally vary for different tracking approaches.
To compensate differences in the skeleton structure, 
we also report results where the offset between ground truth and estimated joint locations is estimated in the first frame and compensated in the remaining frames, reducing the reconstruction error to 3–5\,cm. 
Datasets without ground-truth data cannot be quantitatively evaluated; however, our shape overlap evaluation results suggest that pose estimation is generally accurate.
In summary, pose estimation is reliable, with only occasional failures in Stage I, although the main focus of our work is on the combination with shape estimation.

\inlineheading{Runtime}
In our experiments, runtime scaled linearly with the number of cameras and frames.
Contour-based shape optimization is efficient: it only takes 3 seconds per view, totaling 15 minutes for 50 frames and 6 views on a standard desktop machine.
Skeleton pose estimation is not the main focus of this work and is not optimized; it takes 10 seconds per frame and view, totaling 50 minutes.

{
	
\floatsetup[table]{capposition=top}
	
\begin{table}[t]
\caption{Pose estimation accuracy measured in mm on the \texttt{HumanEva-I} dataset. The standard deviation is reported in parentheses.}
\newcommand{\greyline}{\arrayrulecolor{black!30}\cline{2-7}\arrayrulecolor{black}}
\resizebox{\columnwidth}{!}{
\begin{tabular}{ p{9mm} | l ? c  | c | c | c | c}
	\centering Seq.                                                     & Trained on & Our         & \shortstack{\phantom{g}Amin\phantom{g}\\2013} & \shortstack{\phantom{g}Sigal\phantom{g}\\2012} & \shortstack{Belagiannis\\2014} & \shortstack{Elhayek\\2015} \\
		\Cline{1.1pt}{1-7}
	\multirow{2}{9mm}{\centering S1, \texttt{Walk}} & general  & 74.9 (21.9) & ---                             & 66                           & 68.3                                       & 66.5                                      \\
	\greyline                                                           & HumanEva & 54.6 (24.2) & 54.5                            & ---                               & ---                                        & ---                                       \\
	\Cline{1.1pt}{1-7}
	\multirow{2}{9mm}{\centering S2, \texttt{Box}}  & general  & 59.7 (15.0) & ---                             & ---                               & 62.7                                       & 60.0                                      \\
	\greyline                                                           & HumanEva & 35.1 (19.0) & 47.7                            & ---                               & ---                                        & ---
\end{tabular}
}
\label{table:humaneva}
\end{table}
}

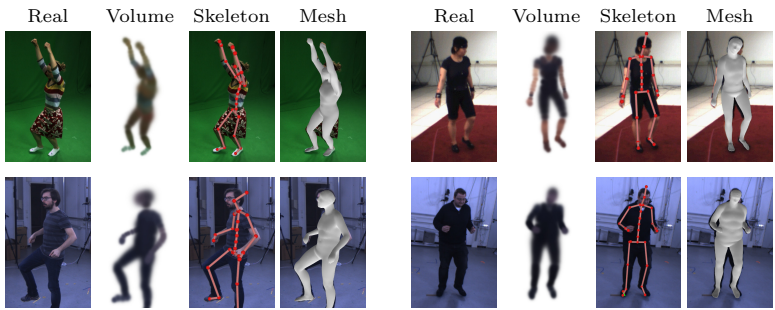
\begin{figure}[t!]
\centering%
		\begin{tikzpicture}[tight background,%
		image/.style={inner sep=0pt},
		subcaption/.style={inner xsep=0.75mm, inner ysep=0.75mm, scale=0.6, above right},]
		\def\padding{2pt}
		\newcommand{\subfig}[3]{\includegraphics[height=0.14\linewidth,angle=0,origin=c,trim=120px 120px 160px 45px,clip]{images/juergen_skirt/#1#2.#3}}
		\newcommand{\subfigB}[3]{\includegraphics[height=0.14\linewidth,angle=0,origin=c,trim=210px 50px 160px 50px,clip]{images/justus1/#1#2.#3}}
		\newcommand{\subfigA}[3]{\includegraphics[height=0.14\linewidth,angle=0,origin=c,trim=135px 0px 235px 100px,clip]{images/dan1/#1#2.#3}}
		\newcommand{\subfigF}[3]{\includegraphics[height=0.14\linewidth,angle=0,origin=c,trim=0px 20px 300px 70px,clip]{images/HumEva_S1_walk/#1#2.#3}}
		\newcommand{\subfigM}[3]{\includegraphics[height=0.14\linewidth,angle=0,origin=c,trim=25px 200px 320px 30px,clip]{images/ms-walk-twoViews/#1#2.#3}}
		\newcommand{\subfigMM}[3]{\includegraphics[height=0.14\linewidth,angle=0,origin=c,trim=320px 200px 50px 30px,clip]{images/ms-walk-twoViews/#1#2.#3}}
				
		\node[image,right=\padding] (f0) at (0,0) {\subfig{Input/image_c0_f}{100}{jpg}};
		\node[image,right=\padding] (f1) at (f0.east) {\subfig{Density/image_c0_f}{100}{jpg}};
		\node[image,right=\padding] (f3) at (f1.east) {\subfig{Skeleton/image_composition_c0_f}{100}{jpg}};
		\node[image,right=\padding] (f4) at (f3.east) {\subfig{Mesh/image_composition_c0_f}{100}{jpg}};
		
		\node[image,right=6mm] (Ff0) at (f4.east) {\subfigF{Input/image_c1_f}{8}{jpg}};
		\node[image,right=\padding] (Ff1) at (Ff0.east) {\subfigF{Density/image_c1_f}{8}{jpg}};
		\node[image,right=\padding] (Ff3) at (Ff1.east) {\subfigF{Skeleton/image_composition_c1_f}{8}{jpg}};
		\node[image,right=\padding] (Ff4) at (Ff3.east) {\subfigF{Mesh/image_composition_c1_f}{8}{jpg}};

		\node[image,below=2mm] (Af0) at (f0.south) {\subfigA{Input/image_c6_f}{700}{jpg}};
		\node[image,below=2mm] (Af1) at (f1.south) {\subfigA{Density/image_c6_f}{700}{jpg}};
		\node[image,below=2mm] (Af3) at (f3.south) {\subfigA{Skeleton/image_composition_c6_f}{700}{jpg}};
		\node[image,below=2mm] (Af4) at (f4.south) {\subfigA{Mesh/image_composition_c6_f}{700}{jpg}};

		\node[image,below=2mm] (Cf0) at (Ff0.south) {\subfigB{Input/image_c5_f}{432}{jpg}};
		\node[image,below=2mm] (Cf1) at (Ff1.south) {\subfigB{Density/image_c5_f}{432}{jpg}};
		\node[image,below=2mm] (Cf3) at (Ff3.south) {\subfigB{Skeleton/image_composition_c5_f}{432}{jpg}};
		\node[image,below=2mm] (Cf4) at (Ff4.south) {\subfigB{Mesh/image_composition_c5_f}{432}{jpg}};
		\node[above=0pt] at (f0.north) {\scriptsize Real};
		\node[above=0pt] at (f1.north) {\scriptsize Volume};
		\node[above=0pt] at (f3.north) {\scriptsize Skeleton};
		\node[above=0pt] at (f4.north) {\scriptsize Mesh};
		
		\node[above=0pt] at (Ff0.north) {\scriptsize Real};
		\node[above=0pt] at (Ff1.north) {\scriptsize Volume};
		\node[above=0pt] at (Ff3.north) {\scriptsize Skeleton};
		\node[above=0pt] at (Ff4.north) {\scriptsize Mesh};
	\end{tikzpicture}
	\caption{In-studio reconstruction of several subjects.
		Our estimates are accurate across diverse body shapes and robust to highly articulated poses.}
	\label{fig:extremeBodyShapes}
\end{figure}

\inlineheading{Limitations and discussion}
Even though the body model was learned from tight clothing scans, our approach handles general apparel well, correctly reconstructing the overall shape and body dimensions. 
We demonstrate that even if not all assumptions are fulfilled, our method produces acceptable results, such as for the dance performance \texttt{Skirt} of Gall et al. \cite{gall2009motion} in \cref{fig:extremeBodyShapes} (top left) that features a skirt.
However, our method was not designed to accurately reconstruct fine wrinkles, facial details, hand articulation, or highly non-rigid clothing.%
We demonstrate fully automatic reconstructions from as few as two cameras and semi-automatic shape estimation using a single image. Fully automatic pose and shape estimates from a single image remains difficult.

\section{Conclusion}
\label{sec:conclusion}

We proposed a fully automatic approach for estimating the shape and pose of a rigged actor model from general multi-view video input with just a few cameras.
\NEW{
It is the first approach that reasons about contours within sum-of-Gaussians representations and which transfers their beneficial properties, such as analytic form and smoothness, and differentiable visibility \cite{rhodin2015iccv}, to the domain of edge- and silhouette-based shape estimation.
This results in an analytic volumetric contour alignment energy that efficiently and fully automatically optimizes the pose and shape parameters.
Based on a new statistical body model,
our approach reconstructs a personalized kinematic skeleton, a volumetric Gaussian density representation with appearance modeling, a surface mesh, and the time-varying poses of an actor.
}
We demonstrated shape estimation and motion capture results on challenging datasets, indoors and outdoors, captured with very few cameras. This is an important step towards making motion capture more practical.

\section*{Acknowledgements}

\NEW{
We thank PerceptiveCode, in particular
Arjun Jain
and
Jonathan Tompson, for providing and installing the ConvNet detector, 
Ahmed Elhayek,
Jürgen Gall,
Peng Guan,
Hansung Kim,
Armin Mustafa and
Leonid Sigal
for providing their data and test sequences, 
The Foundry for license support, and all our actors.
This research was funded by the ERC Starting Grant project CapReal (335545).}

\bibliographystyle{splncs}
\bibliography{610}
\end{document}